\documentclass[journal]{IEEEtran}
\ifCLASSINFOpdf
\else
\fi
\usepackage{graphicx}
\usepackage{float}
\usepackage{subcaption}
\usepackage{amsmath,amssymb,mathtools,bbm}
\usepackage{tipa}
\usepackage{esint}
\usepackage{bbold}
\usepackage{color}
\usepackage{cite}
\usepackage{tikz}
\usetikzlibrary{shapes}
\usepackage{longtable}
\usepackage{pgfplots}
\usepackage{subeqnarray}
\usepackage{dsfont}
\usepackage{changepage}

\usepackage{epsfig} 
\usepackage{mathtools,bm,bbold} 
\usepackage{hyperref} 

\usepackage{algorithmic}
\usepackage{algorithm}

\begin{document}
%
\title{Control of uniflagellar soft robots at low Reynolds number using buckling instability}
%
%
%

\author{Mojtaba Forghani, Weicheng Huang, and M. Khalid Jawed
\thanks{Mojtaba Forghani is with the Department of Mechanical Engineering, Stanford University, Stanford, California 94301 USA (email: mojtaba@stanford.edu).}
\thanks{Weicheng Huang and M. Khalid Jawed are with the Department of Mechanical \& Aerospace Engineering, University of California Los Angeles, Los Angeles, California 90095 USA (email: khalidjm@seas.ucla.edu ).}
}

\maketitle

\begin{abstract} \it
In this paper, we analyze the inverse dynamics and control of a bacteria-inspired uniflagellar robot in a fluid medium at low Reynolds number. Inspired by the mechanism behind the locomotion of flagellated bacteria, we consider a robot comprised of a flagellum---a flexible helical filament---connected to a spherical head. The flagellum rotates about the head at a controlled angular velocity and generates a propulsive force that moves the robot forward. When the angular velocity exceeds a threshold value, the hydrodynamic force exerted by the fluid can cause the soft flagellum to buckle, characterized by a dramatic change in its shape. In this computational study, a fluid-structure interaction model that combines Discrete Elastic Rods (DER) algorithm with Lighthill\rq{}s Slender Body Theory (LSBT) is employed to simulate the locomotion and deformation of the robot. We demonstrate that the robot can follow a prescribed path in three dimensional space by exploiting buckling of the flagellum. The control scheme involves only a single (binary) scalar input---the angular velocity of the flagellum. By triggering the buckling instability at the right moment, the robot can follow the path in three dimensional space. We also show that the complexity of the dynamics of the helical filament can be captured using a deep neural network, from which we identify the input-output functional relationship between the control input and the trajectory of the robot. Furthermore, our study underscores the potential role of buckling in the locomotion of natural bacteria.
\end{abstract}


%
\IEEEpeerreviewmaketitle

\section{Introduction}
\label{intro}
Thanks to the recent advances in MEMS technology, design, fabrication, and control of millimeter and sub-millimeter size robots have received numerous attention with particular emphasis on biomedical applications \cite{Mukherjee_1, Mukherjee_2, app_Ornes, surgery_Taylor, drug_Nelson, prop_small, prop_Nain}. Inspired by the motion of microorganisms in fluid, a number of these works have focused on design of bio-inspired robots for the eventual purpose of placing them inside biological systems to perform non-invasive (or minimally invasive) tasks such as drug/cargo delivery \cite{drug_Li, drug_Nelson, drug_Fusco, cargo_Medina}, surgery \cite{surgery_Edd, surgery_Taylor}, targeted therapy \cite{app_Ornes}, material removal \cite{drug_Nelson}, and imaging \cite{image_Nelson}. The majority of these works are centered around efficient fabrication and propulsion mechanism of micro-robots \cite{prop_Pak, prop_Ghosh, prop_Abbot, prop_small, prop_Nain, prop_Magdanz, prop_Bell}; notable among them are traveling wave propulsion \cite{wave_Behkam, distributed}, chemically powered propulsion \cite{chem}, magnetic field \cite{prop_Ghosh, prop_Nain, drug_Nelson, image_Nelson, prop_small, prop_Abbot, prop_Pak, drug_Li, ctrl_Nour, magnet_Kim, prop_Bell, prop_Magdanz, 2D}, and electric field \cite{electric_Osada}. Others have focused on the characterization of motion dynamics of microorganisms and its relevance to efficient control of micro-robots \cite{drug_Li, ctrl_Nour, ctrl_Lobaton}.

There are a number of limitations associated with the current propulsion mechanisms. For instance, requirement of external magnetic or electric fields and consequently the issues associated with the supply of sufficient field strength (such as the distance between the external source and the robot or the size of the external magnets) \cite{prop_small}, rigidity of the flagellum attached to the robot body (bacteria have flexible flagella) which consequently requires multiple actuators to provide forward motion as well as steering \cite{surgery_Edd}, limited controllability of the robot \cite{RBC}, constraining the robot to operate in two-dimensional (2D) space \cite{2D}, and requiring more than one flagellum (more than 90$\%$ of bacteria are uniflagellar \cite{multi_Nguyen}), not only add to the complexity of the propulsion mechanism, but also leads to inefficient designs (for instance, multiple flagella are not necessarily a means for bacteria to increase their maneuverability). The complexities observed in many of the current designs can be traced back to our unfamiliarity  with the locomotion of microorganisms. This necessitates the search for a mechanism that mimics that of microbial locomotion as close as possible by leveraging the size of the microbes and the medium in which they are operating, e.g., small Reynolds number (ratio of the inertial forces to viscous forces) regime.

Since reciprocal motion mechanisms (deformations with time-reversal symmetry)---such as the ones used by fish---cannot lead to any net propulsion at low Reynolds numbers as a consequence of scallop theorem \cite{prop_Pak}, microorganisms are bound to employ alternative motion mechanisms. Among the different locomotion mechanisms used by microorganisms, propulsion of bacteria by rotation of flagella (flexible helical filaments) has attracted many researchers~\cite{poly_Ali, bundle_Hintsche, buckle_Nguyen, multi_Nguyen, buckle_Son}. By exploiting buckling instability, uniflagellated bacteria alternate their motion between turning and straight path to follow consecutive environmental cues \cite{buckle_Son}. In particular, by choosing the proper angular velocity of the flagellum attached to their body, bacteria can maneuver in three-dimensional (3D) space. 

Inspired by the effectiveness of the motion of flagellated microorganisms, in this work, we have considered a motion mechanism that mimics that of uniflagellated bacteria and proposed a simple strategy for its control. Uniflagellated bacteria-inspired actuation mechanism has significant advantages over other mechanisms. For instance, we will show in this paper that a robot equipped with such mechanism can explore the 3D space using only one input (one rotating actuator). Other actuation methods such as traveling-wave propellers \cite{distributed, wave_Behkam} require distributed actuation which has been proven to be very challenging, in particular at the microscale \cite{drug_Nelson}. Furthermore, depending on the cargo that the robot carries, long strips of drug, shaped into a helical shape, may be used as the flagellum itself, significantly minimizing the weight of the robot \cite{drug_Li}. 

This propulsion method is comparatively more pliable than, for instance, magnetically driven actuators that are limited by rigidity of the material used as the flagellum \cite{image_Nelson} and require specific material properties (such as ferromagnetism) for efficient interaction with the magnetic field \cite{prop_Ghosh, prop_Abbot}. Another advantage of such propellers is their higher maneuverability; due to crawling-like motion of elastic flagellum and flexibility of its dimension \cite{drug_Li}, these robots may potentially be suitable for operating in narrow tubular environments such as moving through lumen \cite{drug_Nelson}. Inspired by these potentials of uniflagellar robots, we attempt to solve the following problem throughout the remainder of this paper: design a control strategy for a uniflagellar robot to follow a prescribed trajectory in three dimensional space, simply by varying the angular velocity of its flagellum.

Due to the complexity of the mechanics of the helical rod and its hydrodynamic interaction with the viscous fluid that it is operating in, numerical methods for simulation of such systems must be able to couple the geometrically nonlinear deformation in the flexible flagellum with the hydrodynamic forces exerted by the fluid. Resistive Force Theory (RFT) is often used as a hydrodynamic model to analyze and predict slender body motion (such as flagella) in fluid \cite{drug_Li, surgery_Edd, prop_Pak, ctrl_Nour}. However, such methods, by neglecting non-local hydrodynamics (such as the interaction among flows generated by distant parts on the flagellum), introduce significant approximations, leading to disagreement between theoretical predictions and experimental observations (see Ref. \cite{rodenborn2013propulsion} for a comparison). Moreover, the mechanical model for the flagellum should allow large deformation, in contrast to the often used assumption of rigid flagellum~\cite{rodenborn2013propulsion} or small deflection~\cite{kim2005deformation}.

In our work, we model the flagellated soft robot as a helical flagellum attached to a rigid spherical head. Here, we ignore the soft hook---the connection between the head and the flagellum in our current study; instead, we mainly focus on the system that comprises a rigid head and a flexible flagellum. The motor is placed inside of the head such that the flagellum can rotate relative to the head and generate the propulsive force for locomotion. As the robot (or the bacterium) moves in the fluid medium, it experiences a hydrodynamic drag force. Due to the microscopic size of bacteria, inertial forces are negligible compared to viscous forces and the fluid flow is at low Reynolds number regime (i.e. $Re \ll 1$). For numerical simulation of this fluid-structure interaction problem, we employ the Discrete Elastic Rods (DER) algorithm \cite{bergou2008discrete, bergou2010discrete, jawed2018primer}---a computational tool for geometrically nonlinear deformation of rod-like structures---in conjunction with Lighthill\rq{}s Slender Body Theory (LSBT) \cite{lighthill1976flagellar}---a model for the hydrodynamic force on rods at low Reynolds number. The accuracy of the said fluid-structure interaction model has been partially validated against experiments in previous works \cite{jawed2015propulsion, jawed2017dynamics, rodenborn2013propulsion}. Furthermore, in our numerical simulator, we have integrated the coupled interaction between the rigid head and the soft filament in low Reynolds environment by balancing the forces and torques in the whole structure \cite{higdon1979hydrodynamic,thawani2018trajectory}. Details behind this model can be found in Ref.~\cite{huang2018numerical}. When the angular velocity of the flagellum is lower than a threshold for buckling, the robot travels along a straight path. However, if the angular velocity exceeds the threshold value, it leads to a buckling instability in the soft filamentary structure and the robot begins to follow a nonlinear trajectory.

Once the proper numerical setup has been established to solve the forward dynamics (the problem of generating robot trajectory from the given angular velocity of its flagellum), we solve the inverse problem (the problem of generating angular velocities of flagellum from a given trajectory) by combining the input that characterizes the path that the robot should follow with a function approximation method (multi-layer neural network) that identifies the functional relationship between the properly projected information of the trajectory and the time-dependent angular velocities. Such inverse problem frameworks have been utilized successfully to control complex dynamical systems where identification of the system, due to blackbox-like behavior of the forward dynamics, is possible only from its input-output observations \cite{robot_inverse}. Exemplary applications include robot-assisted source identification \cite{source_inverse}, stochastic reachable set parameterization \cite{driver_inverse}, robot-human handover tasks \cite{human_inverse}, human response time identification in semi-autonomous systems \cite{reaction_inverse}, and economic dynamics \cite{price_pred}.

The remainder of the paper is organized as follows. In section \ref{problem}, we formulate the problem that we are addressing here. In section \ref{dynamic}, we briefly discuss the dynamics of the helical flagellum. In section \ref{inverse}, we propose a learning process that can be employed to extract the inverse dynamics for the eventual purpose of controlling robot motion. In section \ref{control}, we provide the detail of the control algorithm. In section \ref{result}, we discuss the result of the implementation of the algorithm; and finally, in section \ref{conclusion}, we provide the conclusion and outline our future direction.

\section{Problem Formulation}
\label{problem}

\begin{figure}[t]
\centerline{\psfig{figure=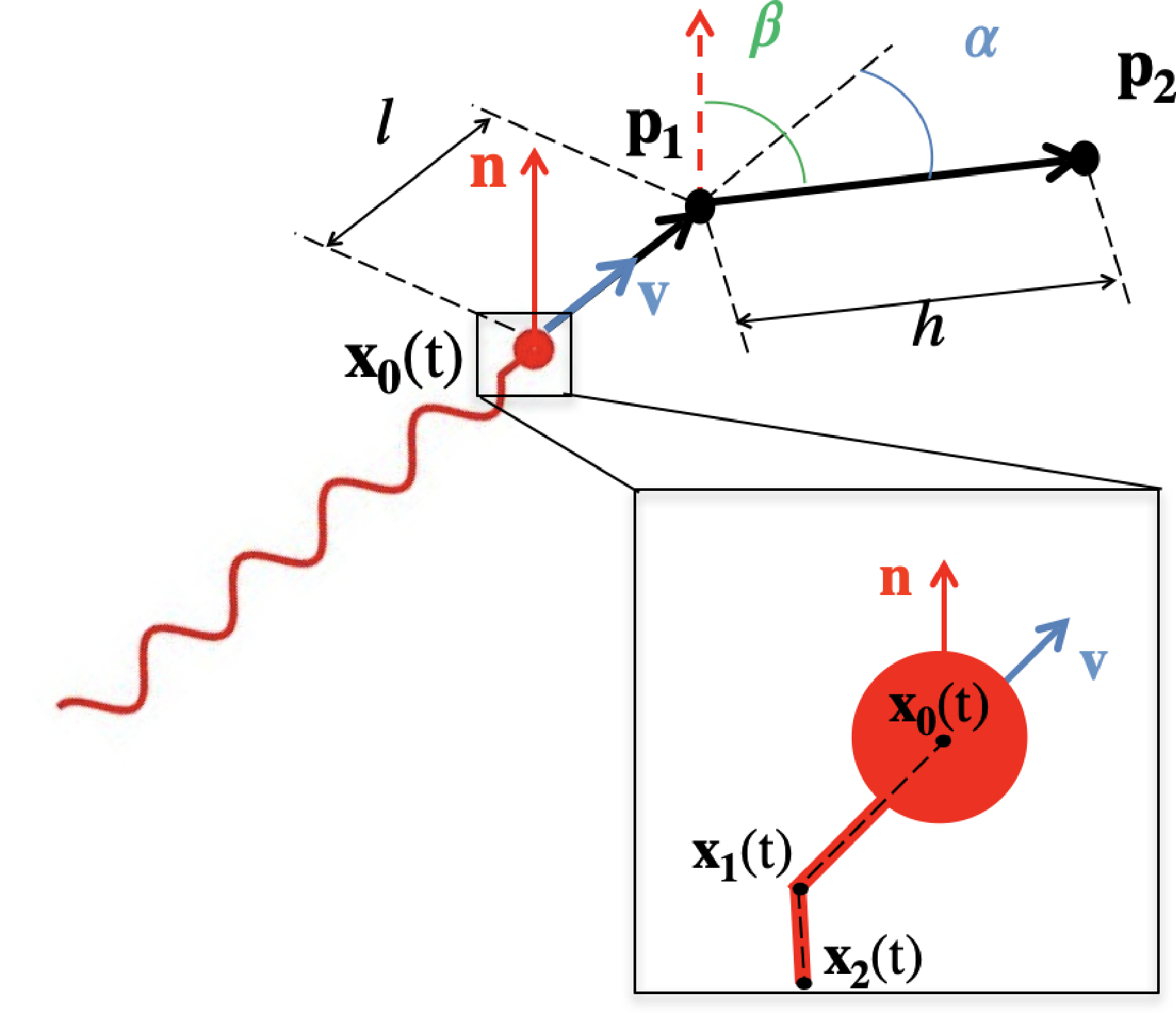,width=2.6in}}
\caption{The robot head (shown as sphere) is currently located at position ${\bf x}_0(t)$ and is moving in the direction of ${\bf v}$. It is expected to follow the trajectory ${\bf x}_0(t) \rightarrow {\bf p}_1 \rightarrow {\bf p}_2$. The points ${\bf p}_1$ and ${\bf p}_2$ are parts of the input at time $t$. Vector ${\bf n}$ determines the orientation of the robot (see Eqn.~(\ref{n_def})).}
\label{sketch}
\end{figure}

Our goal is to provide an algorithm that takes state of the robot (a structure comprising of a head and a soft filament attached to it) and the set of inputs characterizing a desired trajectory (a path that the robot head is expected to follow), and provides proper control input in the form of a time-dependent angular velocity. The characterization of the trajectory consists of two points in space that the robot is expected to cover in its upcoming time steps. Any curve in the 3D space can be discretized into linear segments, allowing implementation of the algorithm as close as possible on arbitrary trajectories. Figure~\ref{sketch} shows a sketch of the problem that we are addressing here.

The input to the control algorithm at time $t$ consists of the current state of the robot and the trajectory that it should follow: 
\begin{equation}
{\bf s} (t)= \left( {\bf x}_0(t- (0:k)\delta t), {\bf x}_{1:2}(t), {\bf p}_{1:2}, \omega(t) \right)^T ,
\label{state}
\end{equation} 
where $i:j= \{i, i+1, ..., j\}$, $\delta t$ is the time interval between two consecutive observations of the position of the system (the robot head), ${\bf x}_i(t)$ is the position of the $i$-th node of the flagellum with ${\bf x}_0(t)$ showing position of the head of the robot (${\bf x}_{0:2}(t)$ are three non-collinear points on the robot determining its orientation; see section~\ref{dynamic} for further detail of the node definition), ${\bf p}_1$ and ${\bf p}_2$ are the two successive points that the robot is expected to follow, $\omega(t)$ is the current angular velocity (the known input to the actuator), and the $T$ superscript stands for transposition. Here, the information of the position of the head node of the $k$ previous time steps, ${\bf x}_0(t- (1:k)\delta t)$, has been used in order to determine the direction of the motion (vector ${\bf v}$ in Fig.~\ref{sketch}) and in order to check whether the trajectory is linear or not. The information of the position of the two most adjacent nodes to the head has been used to determine the orientation of the robot (the unit vector ${\bf n}$ in Fig.~\ref{sketch}). If the orientation is measurable through other sensory information, ${\bf x}_{1:2}(t)$ in Eqn.~(\ref{state}) can be replaced with ${\bf n}$. The orientation (${\bf n}$) is related to the direction of motion (${\bf v}$) and ${\bf x}_{1:2}(t)$ through the following relationship
\begin{equation}
{\bf n}= \frac{{\bf v} \times \left( {\bf x}_1(t)- {\bf x}_2(t) \right)}{\big|\big| {\bf v} \times \left( {\bf x}_1(t)- {\bf x}_2(t) \right) \big|\big|} .
\label{n_def}
\end{equation}
More detail of this parameter and its relation to the control of the robot trajectory is provided in section \ref{control}.

The control policy determined at time $t$ that characterizes the future trajectory of the robot (for time $t_f > t$) amounts to determining the time dependent angular velocity $\Omega \left(t_f, {\bf s}(t) \right)$. The time $t_f$ (final time) is the time that it takes for the robot to reach point ${\bf p}_2$ in space from its current location (${\bf x}_0(t)$). Therefore, it is not {\it a priori} known; since the translational velocity of the robot is constrained (as we will explain in further detail in sections \ref{inverse} and \ref{control}).

\begin{figure}[t]
\centerline{\psfig{figure=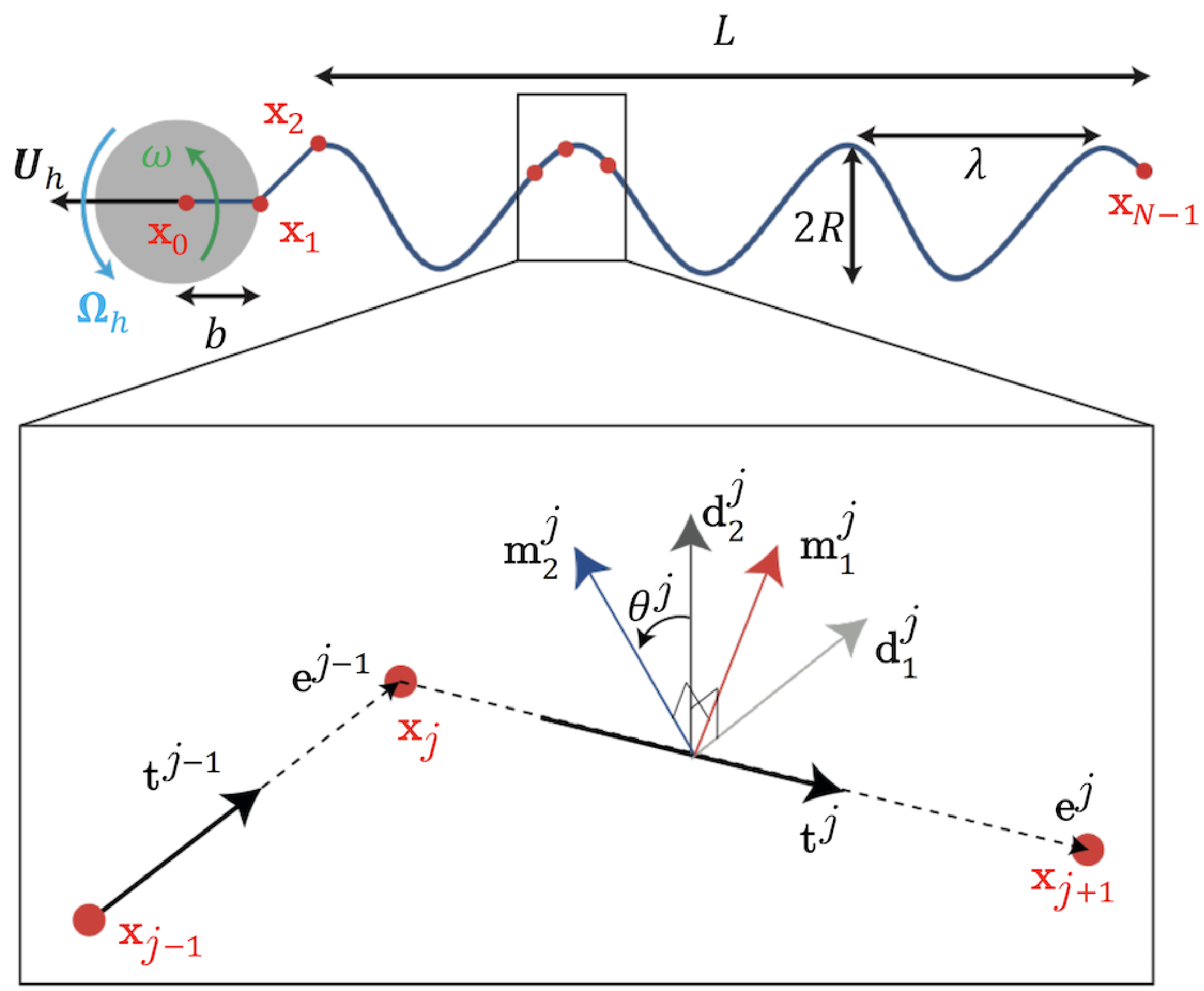,width=3.2in}}
\caption{The geometry of the flagellated soft robot and the interaction between its rigid head and soft filament. Inset: relevant quantities of the discrete rod.}
\label{Geometry}
\end{figure}

\paragraph*{Physical Parameters}
In this fluid-structure interaction problem, the structure is comprised of a Kirchhoff elastic rod~\cite{kirchhoff1859uber}---our model for the flagellum---attached to a rigid head at one end. Based on prior experiments~\cite{jawed2015propulsion} on the dynamics of soft helical rods in viscous fluid, the physical parameters of the right-handed helical flagellum shown schematically in Fig.~\ref{Geometry} are: axial length $L=130$ mm, pitch $\lambda = 32.6$ mm, helix radius $R=6.04$ mm, (within the biological regime~\cite{rodenborn2013propulsion}), radius of circular cross-section $r_0 = 1$ mm, Young\rq{}s modulus $E=1$ MPa, and Poisson\rq{}s ratio $\nu = 0.5$ (incompressible material). The head is spherical with a radius $b=10$ mm. The fluid medium (assumed to be glycerin) has a viscosity of $\mu = 2.7$ Pa$\cdot$s. The density of the rod is $\rho = 1.27$ g/cm$^3$ (here, the rod density is enlarged 100 times for computational efficiency in the forward physics-based simulation~\cite{jawed2017dynamics}) such that the Reynolds number $Re$ remains in Stokes regime. While this density value does not affect the dynamics as long as we maintain low Reynolds number, a larger value of density typically allows us to take larger time step, $\Delta t$, in the simulation and reduces the computation time. For this study, we chose $\Delta t = 1$ ms. These parameters are similar to the ones used in our prior study~\cite{huang2018numerical}. The physical system would not change as long as the non-dimensional parameters remain unchanged~\cite{huang2018numerical}. The dimensionless sperm number in our system is given by $\omega \eta_{\perp} L^4 / EI$, where $\eta_{\perp} = 4 \pi \mu / \log(2 \lambda / r_0 + 0.5)$. Here, the sperm number is $\{359, 1580 \}$ ~\cite{coq2008rotational}.

\section{Forward Dynamics}
\label{dynamic}

In this section, we provide an overview of the simulation procedure of the dynamics of the robot at low Reynolds number. Detailed description of the numerical method used for the forward dynamics is provided in Ref. \cite{huang2018numerical}. As shown in Fig.~\ref{Geometry} (Inset), the rod in DER~\cite{bergou2010discrete, jawed2018primer} is discretized into $N$ nodes (here, $N=122$, similar to~\cite{huang2018numerical}) located at $\mathbf x_j$ with $j=0, \ldots, N-1$. The segment of the rod between two nodes, $\mathbf x_{j}$ and $\mathbf x_{j+1}$, is the edge vector $\mathbf e^j = \mathbf x_{j+1} - \mathbf x_j$. We use subscripts for node-based quantities and superscripts for edge-based quantities. Associated with each of $N-1$ edges is the orthonormal material frame $\left( \mathbf m^j_1, \mathbf m^j_2, \mathbf t^j \right)$ that stays adapted to the centerline, i.e. $\mathbf t^j$ is the unit tangent vector along $\mathbf e^j$. A second orthonormal adapted frame $\left( \mathbf d^j_1, \mathbf d^j_2, \mathbf t^j \right)$ is used as the reference frame that stays adapted to the centerline through {\em parallel transport in time}~\cite{jawed2018primer}. The relative angular orientation between the reference frame and the material frame at each edge is the twist angle $\theta^j$ with $j=0, \ldots, N-2$. The nodal coordinates, $\mathbf x_j$, and the twist angles, $\theta^j$, constitute the $4N-1$ sized degrees of freedom (DOF) vector, $\mathbf{q} = [ \mathbf x_0, \theta^0, \ldots, \mathbf x_{N-2}, \theta^{N-2}, \mathbf x_{N-1} ]$, that completely describes the configuration of the rod. The DER algorithm marches forward in discrete time steps and updates $\mathbf q$ based on the balance of forces at each DOF. The equation of motion at $i$-th DOF, $q_i$, ($i=0, \ldots, 4N-2$) to move from time $t$ to $t + \Delta t$ is 
\begin{multline}
m_i \frac{q_i (t + \Delta t) - q_i (t) } {\Delta t^{2}} - 
m_i \frac{ \dot{q}_i (t)} {\Delta t} - \\
{f}_{i}^\text{int} \left( \mathbf q (t + \Delta t) \right)
- {f}_{i}^\text{ext} \left( \mathbf q (t) \right) = 0,
\label{EOM}
\end{multline}
where $m_i$ is the lumped mass at $q_i$, $\dot{q}_i (t)$ is the velocity at time $t$ (i.e. time derivative of the DOF), $f_{i}^\text{int}  \left( \mathbf q (t + \Delta t) \right)$ is the sum of elastic stretching, twisting, and bending forces that can be evaluated from $\mathbf q (t + \Delta t)$ (details can be found in Ref.~\cite{bergou2010discrete}), and $f_{i}^\text{ext} \left( \mathbf q (t) \right)$ is the external hydrodynamic force computed from $\mathbf q (t)$, described later in this section. When $q_i (t)$ and $\dot{q}_i (t)$ are known, the system of $4N-1$ equations in Eqn.~(\ref{EOM}) can be solved using Newton\rq{}s method to obtain $q_i (t + \Delta t)$. The velocity at $t+\Delta t$ is simply $\dot{q}_i (t + \Delta t) = \left(q_i (t + \Delta t) - q ( t ) \right) / \Delta t$.

In order to incorporate the rigid head in this framework, the first node, $\mathbf x_0$, is assumed to be the center of the head, therefore, the velocity of the head is simply $\mathbf U_h = \dot {\mathbf x}_0$. The joint (the connection between the head and the flagellum) is the first edge of the flagellum and is assumed to be rigid in our framework. As shown in Fig.~\ref{Geometry}, the second node, $\mathbf x_1$, is located on the axis of the helix such that the first edge $\mathbf e^0 = \mathbf x_1 - \mathbf x_0$ is parallel to this axis. The flagellum rotates at a prescribed angular speed, $\omega$, along its first edge, $\mathbf e^0$---the control parameter of this problem. When $\omega$ is positive, the flagellum rotates counter clockwise as viewed from above the head. The torque generated by the hydrodynamic force on the flagellum causes the head to rotate at an angular velocity $\bm{ \Omega}_h$. In the real physical system, the angular velocity of the motor is the sum of the angular velocity from the flagellum and the head, $|| \omega || + || \bm{ \Omega}_h ||$.

As the flagellum and the head move in the fluid, they experience an external hydrodynamic force, represented by ${f}_{i}^\text{ext}$ in Eqn.~(\ref{EOM}). The motion of the flagellum influences the force exerted by the fluid on the head; reciprocally, the motion of the head affects the force on the flagellum, resulting in a highly coupled problem, as summarized next.

\paragraph*{Force on the Flagellum} 
The velocity at each node on the rod is equal to the fluid velocity at that point (no-slip boundary condition). The velocity at the $j$-th node on the flagellum, $\mathbf u_j \equiv \left[ \dot q_{4j}, \, \dot q_{4j+1}, \, \dot q_{4j+2} \right]$, can be decomposed into two components: (i) a flow $\left(\mathbf{u}_{f} \right)_j$ that is generated by the force exerted by the flagellum onto the fluid (equal and opposite of the hydrodynamic force on the flagellum), and (ii) another flow $\left(\mathbf{u}_{h} \right)_j$ that is induced by the motion of the head. For the first component, we use LSBT which relates the velocity, $\left(\mathbf{u}_{f} \right)_j$, and the hydrodynamic force on the flagellum~\cite{jawed2015propulsion, lighthill1976flagellar},
\begin{equation}
- \left(\mathbf{u}_{f} \right)_j = \frac { \left(\mathbf{f}_{j} \right)_{\perp} } { 4 \pi \mu (2 \delta) } +
\sum_{k=1, k \neq j}^{N-1}
\frac {1} {8 \pi \mu || \mathbf{r}_{jk} ||}
\left[
\mathbb{I} + \mathbf{\hat r}_{jk} \otimes \mathbf{\hat r}_{jk}
\right] \mathbf{f}_k,
\label{LighthillSBT}
\end{equation}
where 
$\mathbf f_j \equiv \left[ f^{\textrm{ext}}_{4j}, \, f^{\textrm{ext}}_{4j+1}, \, f^{\textrm{ext}}_{4j+2} \right]$ is the external force at the $j$-th node, 
$\left(\mathbf{f}_{j} \right)_{\perp} = \mathbf{f}_{j} \cdot \left( \mathbb{I} - \mathbf{t}^j \otimes \mathbf{t}^j \right)$ is the projection of $\mathbf{f}_{j}$ along the tangent $\mathbf{t}^j$ at that node, $\mathbb{I}$ is the identity tensor, 
``$\cdot$" denotes the dot product,  $\otimes$ notation is the tensor product, 
$||...||$ is the norm-2 of a vector, $\mathbf{r}_{jk}$ is the position vector from $k$-th to $j$-th node, $\mathbf{\hat r}_{jk}$ is the unit vector along $\mathbf{r}_{jk}$, and $\delta = r_0 \sqrt e/2$ is the natural cutoff length ($r_0$ is the radius of the circular cross-section of the rod and $e$ is the Napier's constant). This discrete formulation of LSBT requires that the length of each edge be $ 2 \delta$. 

The moving head with the translational velocity $ \mathbf{U}_{h} $ and angular velocity $ \bm{\Omega}_{h} $ also contributes to the flow along the flagellar filament, leading to the following component \cite{higdon1979hydrodynamic,thawani2018trajectory}
\begin{multline}
\left(\mathbf{u}_{h} \right)_j = \frac {b^3} {(r_h)_j^3} (\mathbf{r}_{h})_j \times \bm{\Omega}_{h}+ \frac {3} {4} b \left[ \left( \frac {\mathbb{I}} {(r_{h})_j} + 
\frac {(\mathbf{r}_{h})_j \otimes (\mathbf{r}_h)_j } {(r_h)_j^3} \right) \right. \\ \left. +\frac {b^2} {3} \left( \frac {\mathbb{I}} {(r_{h})_j^3} - \frac {(\mathbf{r}_{h})_j \otimes (\mathbf{r}_{h})_j} {(r_{h})_j^5} \right) \right] \cdot \mathbf{U}_{h},
\label{velocityHead}
\end{multline}
where $(\mathbf{r}_{h})_j$ is the position vector of the $j$-th node relative to the center of the head, $(r_h)_j= ||(\mathbf{r}_h)_j||$, and $\times$ notation is the cross product. Combining Eqns.~(\ref{LighthillSBT}) and (\ref{velocityHead}), the actual velocity at the $j$-th node is $\mathbf u_j = \left(\mathbf{u}_{f} \right)_j + \left(\mathbf{u}_{h} \right)_j$.

\paragraph*{Force on the Head}
The flow around the head is influenced by both the moving fluid current around it and the motion of the flagellum. The viscous fluid generates the drag force $ -6\pi \mu b \mathbf{U}_{h} $ and torque $ -8\pi \mu b^3 \bm{\Omega}_{h} $ on the moving head, while the flow caused by the force at each node on the flagellum (see Eqn.~(\ref{LighthillSBT})) results in the following force and torque on the head~\cite{higdon1979hydrodynamic,thawani2018trajectory}:
\begin{align}
\mathbf{f}_{h} & = \sum_{j = 1}^{N -1} \left(- \frac {3} {2} \frac {b} {(r_{h})_j } + \frac {1} {2} \frac {b^3} {(r_{h})_j^3} \right) \mathbf{f}_{j} + \nonumber \\ & \hspace{55pt} \frac {\mathbf{f}_{j} \cdot (\mathbf{r}_{h})_j } {(r_{h})_j^2} \left(- \frac {3} {4} \frac {b} {(r_{h})_j } + \frac {3} {4} \frac {b^3} {(r_{h})_j^3} \right)(\mathbf{r}_h)_j, \nonumber \\
\mathbf{t}_{h} & = \sum_{j = 1}^{N -1} - \frac{b^3} {(r_h)_j^3} (\mathbf{r}_{h})_j \times \mathbf{f}_{j}.
\label{ForceHead}
\end{align}
Specifically, these equations are derived from the Faxen’s law.

Here, we briefly review the scheme to compute the $4N-1$ sized external force vector (represented by $f^{\textrm{ext}}_i$ in Eqns.~(\ref{EOM})). At each time step of the simulation (between time $t$ to $t+\Delta t$), the DOF vector, $\mathbf q$, the velocity, $\dot{ \mathbf q}$, the angular velocity of the first edge on flagellum, $\omega$, and the angular velocity of the head, $ \bm{\Omega}_{h} $, from the previous time step, $t$, are known. We use Eqn.~(\ref{velocityHead}) to evaluate the flow caused by the moving head, $\left(\mathbf{u}_{h} \right)_j$. Then the flow generated by the flagellum is $\left(\mathbf{u}_{f} \right)_j = \mathbf{u}_j - \left(\mathbf{u}_{h} \right)_j$. Equation~(\ref{LighthillSBT}) can then be solved to compute the external force at any $j$-th node, $\mathbf f_j  \equiv \left[ f^{\textrm{ext}}_{4j}, \, f^{\textrm{ext}}_{4j+1}, \, f^{\textrm{ext}}_{4j+2} \right]$, where $j = 0, \ldots, N-1$. Once $\mathbf f_j$ is calculated, Eqn.~(\ref{EOM}) can be used to obtain the new DOF vector and velocity at time $t+\Delta t$, treating the elastic forces implicitly and external forces explicitly. Lastly, the condition of torque balance on the whole robot is used to obtain the angular velocity of the rotating head, $ \bm{\Omega}_{h} $.

In summary, the input of our numerical tool is the control parameter $\omega (t)$ that can vary as a function of time, and the output is the configuration of the rod (e.g. the DOF vector, $\mathbf q (t)$) that evolves with time. In this study, we are primarily concerned with the coordinates of the first node (i.e. $\mathbf x_0$---the first three elements of the DOF vector) that represents the trajectory of the head.

\begin{figure}[t]
\centerline{\psfig{figure=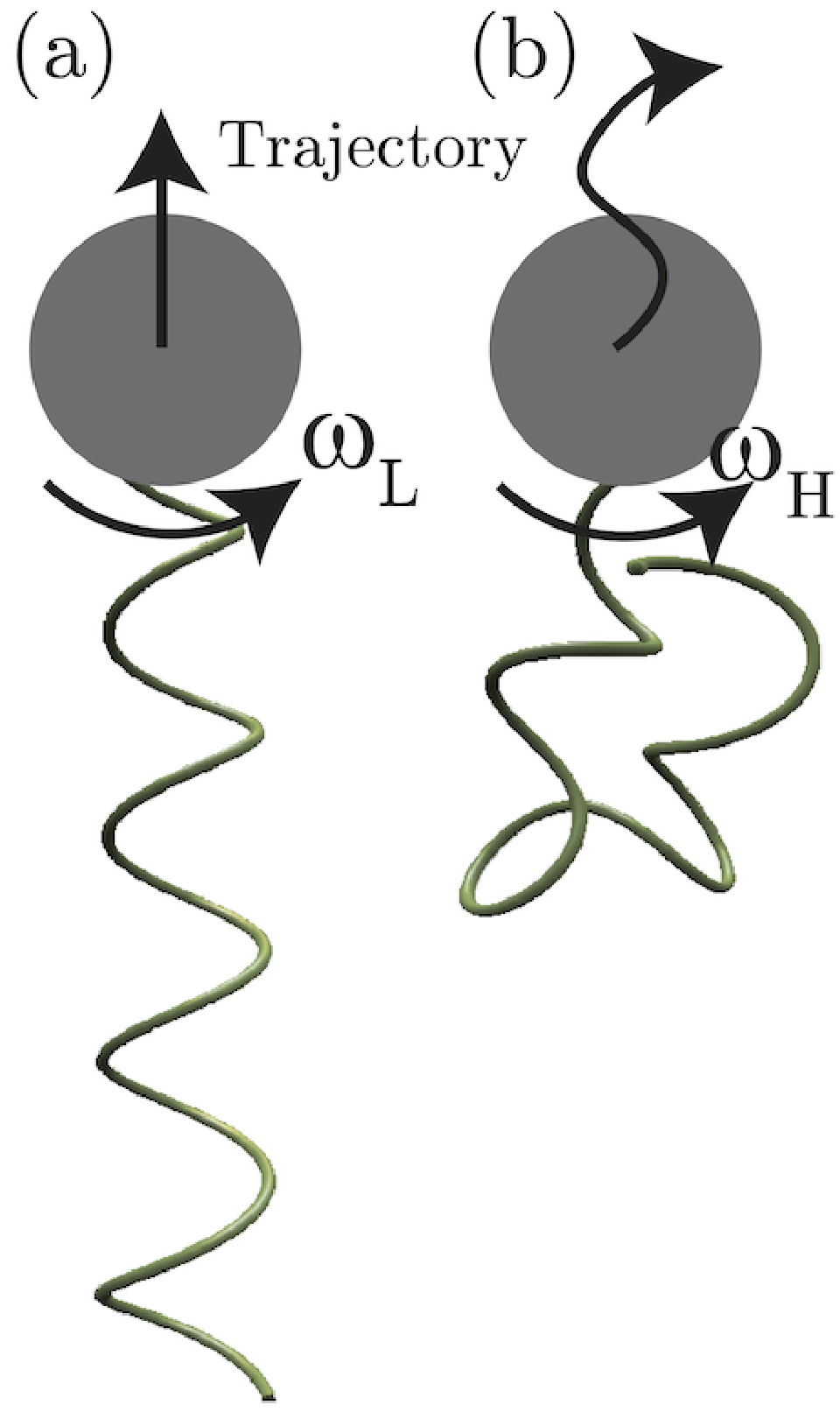,width=1.5in}}
\caption{Schematic of (a) an unbuckled flagellum at $\omega (t) = \omega_L < \omega_b$ and (b) a buckled flagellum at $\omega (t) = \omega_H > \omega_b$.}
\label{buckling}
\end{figure}

A key feature of the dynamics of the robot is the presence of a buckling instability~\cite{vogel2012motor, jawed2015propulsion}. In Fig.~\ref{buckling}(a), when the angular velocity $\omega$ stays below the threshold angular velocity for buckling, $\omega_b$, the flagellum retains its helical shape and moves along a straight line parallel to the axis of the helix. The translation distance is linearly related to the angular velocity in this phase. However, as shown schematically in Fig.~\ref{buckling}(b), if $\omega$ exceeds the threshold $\omega_b$, the excessive hydrodynamic drag can cause the flagellum to buckle and a nonlinear trajectory ensues, in which the turning angle is a function of the running time of the higher angular velocity ~\cite{huang2018numerical}. For the model robot considered in our study, the threshold buckling velocity is $\omega_b \approx 5.8$ rpm based on our previous non-dimensional study~\cite{huang2018numerical}.

\section{Inverse Dynamics}
\label{inverse}

The goal here is to find the functional relationship between a set of inputs characterizing a given path and the time dependent angular velocities. In particular, we seek to find the functions $t_H= f_H (h, \alpha)$, $t_L= f_L (h, \alpha)$, $\beta= f_\beta\left( t_H, t_L \right)$, and $l= f_l\left( t_H, t_L \right)$ such that the input
\begin{align}
\omega(t; t_0, t_H, t_L) &= \begin{cases}
\omega_L & \text{if } t- t_0 < 0,\\
\omega_H & \text{if } 0 \leq t- t_0 < t_H, \\ 
\omega_L & \text{if } t_H \leq t- t_0 < t_H+ t_L, 
\end{cases}
\label{omega_learn}
\end{align}
the angular velocity parameterized with $t_0, t_H$, and $t_L$, generates a trajectory ${\bf x}_0( t_0\leq t< t_0+ t_H+ t_L)$ that is parameterized with $h, \alpha, \beta$, and $l$, in which
\begin{align}
\alpha &= \frac{180}{\pi}\cos^{-1} \left[ \frac{ \left( {\bf p}_2- {\bf p}_1 \right) \cdot {\bf v} }{ \big|\big| {\bf p}_2- {\bf p}_1 \big|\big| } \right] , \nonumber \\
\beta &= \frac{180}{\pi} \tan^{-1} \left[ \frac{ \left( {\bf p}_2- {\bf p}_1 \right) \cdot \left( {\bf v} \times {\bf n} \right) }{ \left( {\bf p}_2- {\bf p}_1 \right) \cdot {\bf n} \big|\big| {\bf v} \times {\bf n} \big|\big| } \right] ,
\label{alphabeta}
\end{align}
characterize the angles (in degrees units) between after-steering linear trajectory and the robot orientation, where ${\bf v}$ is the unit vector defining the direction of motion (see Fig. \ref{sketch}).

In Eqn.~(\ref{omega_learn}), $\omega_L$ is an angular velocity below the flagellum buckling ($\omega_L < \omega_b$) for which the path of the robot can be well approximated by a line, while $\omega_H$ is an angular velocity above the buckling ($\omega_H > \omega_b$) for which the robot follows a complex path in 3D which is responsible for steering of the robot. The time $t_0 > 0$ is a small time interval chosen such that the characterization of the trajectory starts when sufficient time has passed since the start of the motion (time 0) to ensure that the non-linearity due to starting from the static configuration has vanished. Time $t_H$ is the time interval in which the high angular velocity, $\omega_H$, is being applied, while time $t_L$ is the time interval in which the low angular velocity, $\omega_L$, is being applied in order to bring the robot trajectory back to linear regime after it has changed the direction. The reason for the choice of functional relationships considered here is that due to the cyclic motion of the robot around its own axis (generating different values of $\beta$), any desired value of $\beta$ can be achieved with negligible compromise on the values of $h$ or $\alpha$, consequently, $\beta$ can be determined via $f_\beta$ function after calculating the required $t_H$ and $t_L$ that satisfy the desired values of $h$ and $\alpha$; further detail is provided in section \ref{control}.

\begin{figure}[t]
\centerline{\psfig{figure=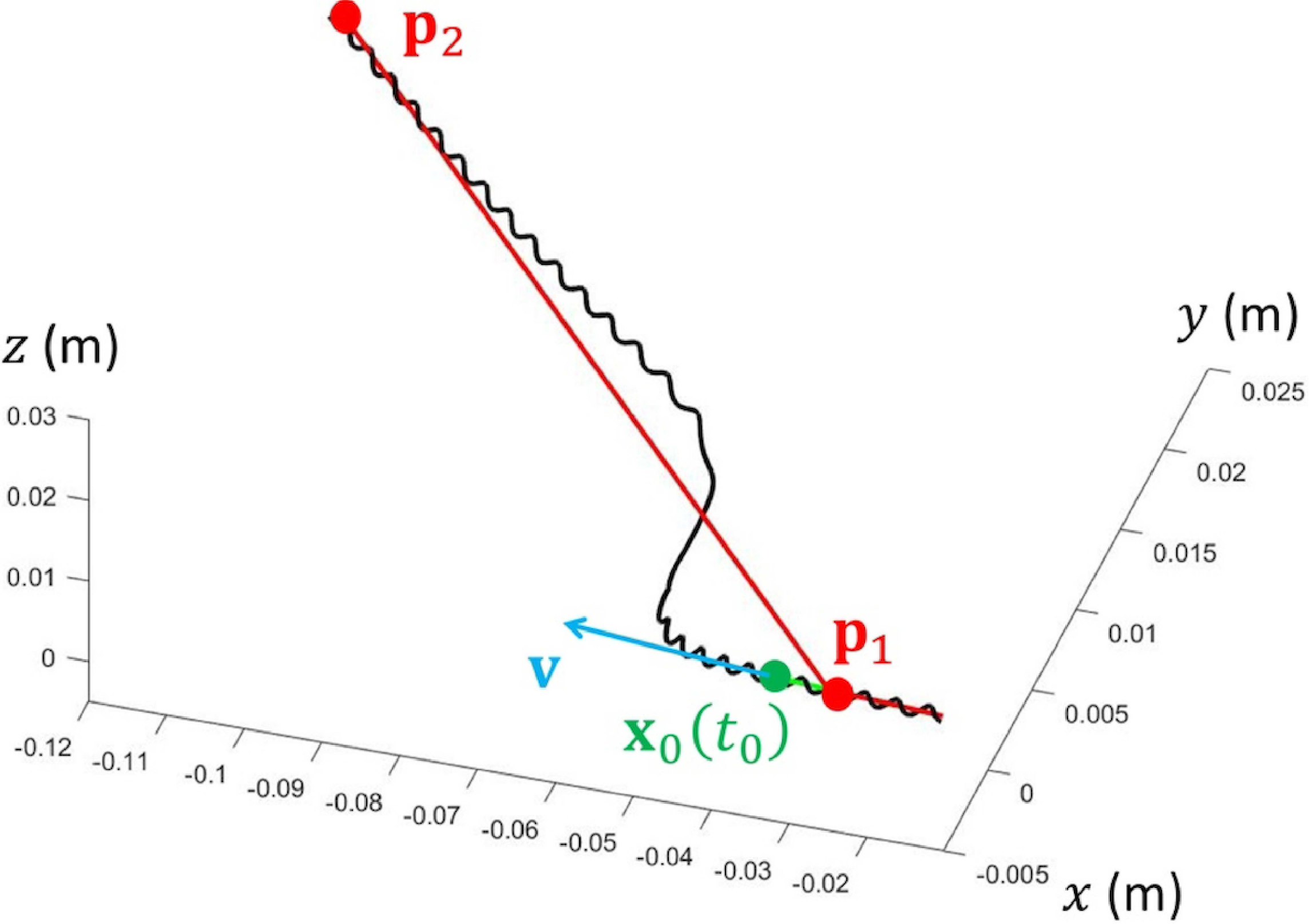,width=3.2in}}
\caption{Parameterization of generated trajectories. The two straight lines are the before- and after- turning linear trajectories based on $h, \alpha, \beta$, and $l$ parameterization, while the helical trajectory is the actual robot trajectory. The robot head (flagellum not shown in the figure) is located at ${\bf x}_0(t_0)$ and at this time ($t_0$) $\omega_H$ is applied. The robot has already passed the point ${\bf p}_1$, meaning that in this figure $l< 0$ (since $\left( {\bf p}_1- {\bf x}_0(t_0) \right) \cdot {\bf v} <0$). The steering happens in a length scale significantly smaller than the robot size, therefore, its non-linearity does not cause significant deviation from ${\bf p}_2- {\bf p}_1$ vector (compare the scales in the figure with length of the flagellum, 200 mm).}
\label{sketch_2}
\end{figure}

The solution to the inverse dynamics problem begins with generation of trajectories with inputs in the form of Eqn.~(\ref{omega_learn}). These trajectories can either be generated incrementally (by increasing $t_H$ and $t_L$ gradually) or randomly (by generating random $t_H$ and $t_L$). Here, we have generated 55 long trajectories (the total time $t_\text{total}= 2500$ s) with different $t_H$ (gradually increasing from 0 to 55 s with 1 s interval), and for each long trajectory, we have picked different segments in the form $t_0\leq t < t_e$, where $t_0+ t_H+ k\delta t< t_e \leq t_\text{total}$ ($t_e$ is the end point), as one datapoint (one dataset in the form $(t_H, t_L, h, \alpha, \beta, l)$). This means that we simulate the forward dynamics based on a long simulation with $\omega(t; t_0, t_H, t_\text{total}- t_H- t_0)$ input (see Eqn.~(\ref{omega_learn})) and then extract the information of $(t_H, t_L, h, \alpha, \beta, l)$ for different segments corresponding to $\omega(t; t_0, t_H, t_L)$ input, where $k\delta t< t_L \leq t_\text{total}- t_H- t_0$, as one datapoint. Note that this process is necessary since the parameters $\alpha$ and $\beta$ change as the robot moves along a long trajectory. Since the trajectory does not become linear immediately after time $t_0+ t_H$, due to the non-linearities associated with the above-buckling regime, and even after reaching linear regime, the two linear trajectories before and after steering are not coplanar, the variation in parameters $\alpha$ and $\beta$ for a given long trajectory, albeit small, is non-zero for the interval $t_0+ t_H+ k\delta t <t < t_\text{total}$. Moreover, even in the linear regime, the trajectory in reality does not follow a linear path; in fact, the trajectory is a helical path with a small radius that has a non-varying axis of rotation (see Fig. \ref{sketch_2}). 

The time $t_0$ here is taken to be 100 s. This is to ensure that the initial non-linearity associated with starting from the static state does not influence the data. Each long trajectory is divided into about 2300 datapoints (the set $(t_H, t_L, h, \alpha, \beta, l)$). This is because the first 100 s of any of the long trajectories are not included in the data (since $t_0= 100$ s), and moreover, we allow, depending on the value of $t_H$, an average of 100 s to pass from $t_0$ in order to ensure when a trajectory reaches its endpoint (time $t_e$ or point ${\bf p}_2$ in Fig. \ref{sketch}), it is in the linear regime. In other words, $t_H+ k\delta t$ is, on average, about 100 s, and can be either smaller or larger than this value depending on $t_H$.  This leads to 2300 (=2500-100-100) datapoints---note that the data is collected from each long trajectory with 1 s intervals. Finally, since we have generated 55 long trajectories, corresponding to $0< t_H <55$ s with 1 s interval, we have been able to generate a dataset of size $2300\times 55 \approx 120000$. Note that the maximum value of 55 s for $t_H$ is due to the fact that at $t_H= 55$ s, the angle $\alpha$ (see Fig.~\ref{sketch}) reaches 90 degrees, implying that $0< t_H <55$ s can cover $0< \alpha <90$ degrees of rotation. We did not see any improvement using a higher resolution of $t_H$ (0.2 s interval instead of 1 s interval, as being used here).   

After generation of sufficient datapoints (different segments of long trajectories), we proceed to convert the information contained in ${\bf x}_0(t_0 \leq t < t_0+ t_H+ t_L)$ to the form $(t_H, t_L, h, \alpha, \beta, l)$. Since the linear trajectories before and after turning are not coplanar, this is done by finding a point ${\bf p}_1$ located on the unique plane defined by ${\bf p}_2= {\bf x}_0(t_0+t_H+ t_L)$ and ${\bf x}_0(t_0)$, that contains ${\bf v}$, the unit vector defining the direction of motion (see Fig.~\ref{sketch_2}). This ensures that the two lines characterizing the steering are coplanar and as close as possible to the true trajectory. The details of finding the optimal ${\bf p}_1$ are provided in Appendix A. 

\begin{figure}[t]
\centerline{\psfig{figure=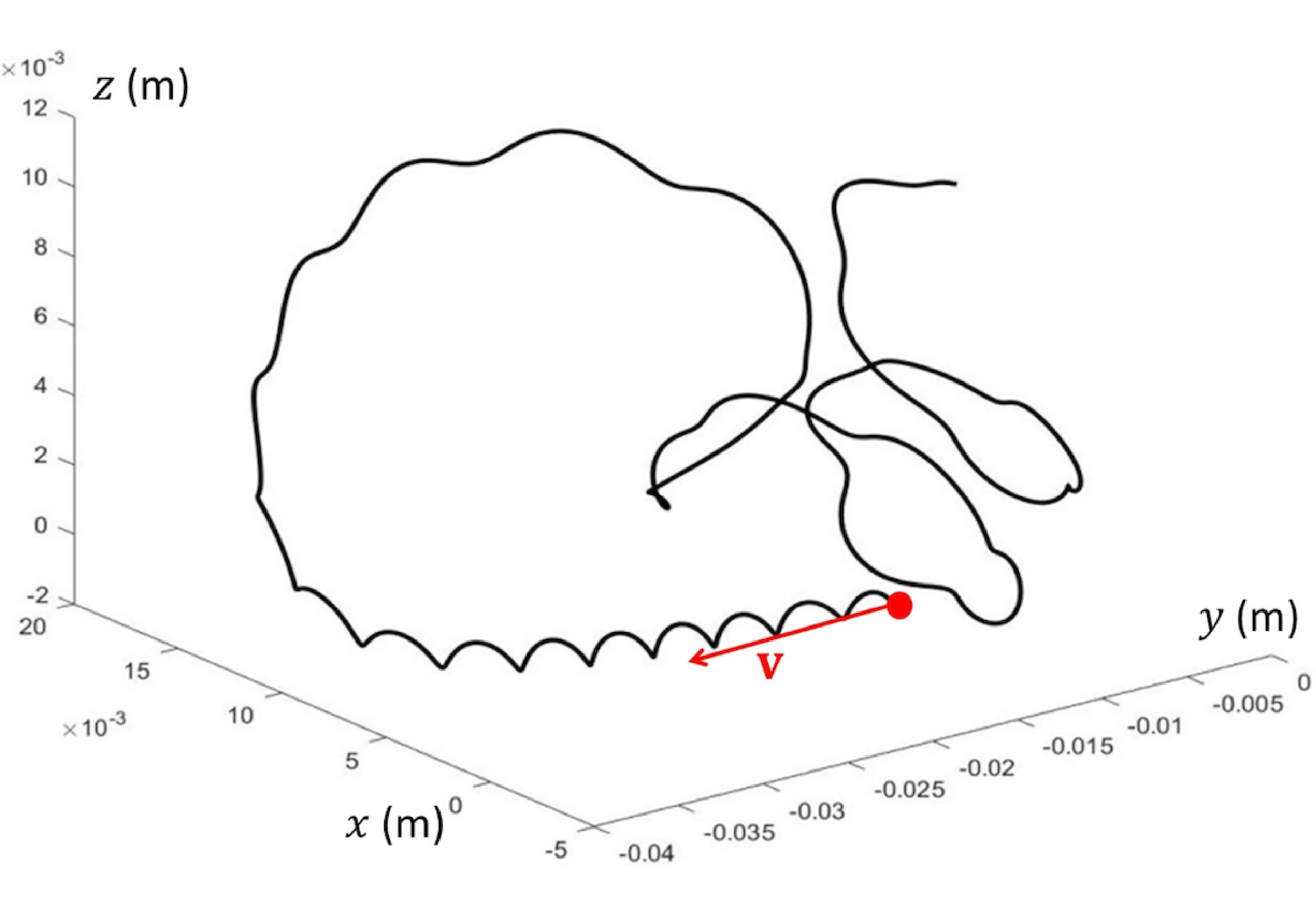,width=3.0in}}
\caption{The trajectory of robot head (flagellum not shown in the figure) when $\omega_H$ is applied. The ${\bf v}$ vector shows the direction of robot when the motion starts.}
\label{w_H_traj}
\end{figure}

\begin{figure}[t]
\centerline{\psfig{figure=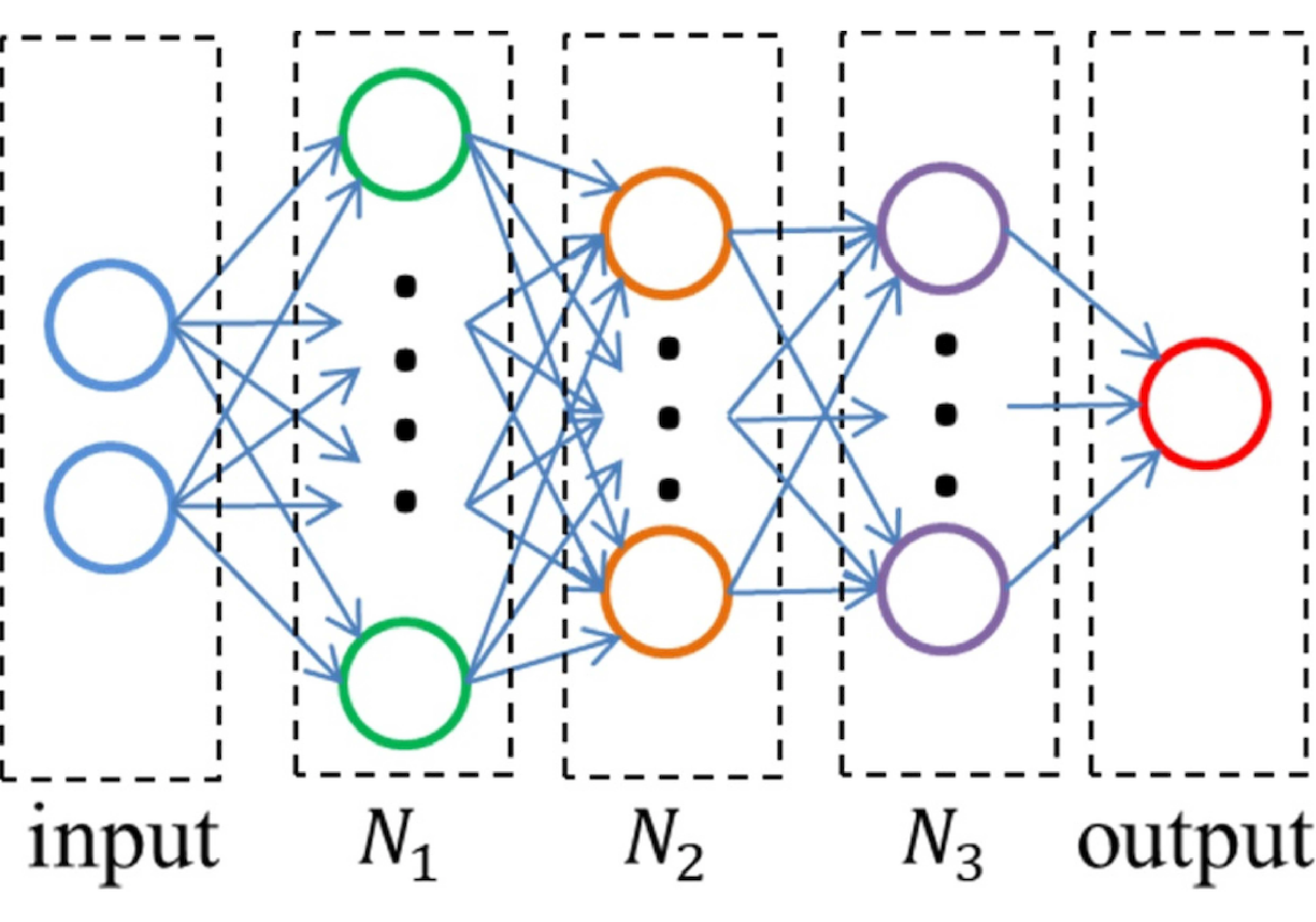,width=2.0in}}
\caption{Neural network used to identify the functions. Here, we used $N_1= 20$, $N_2= 10$, and $N_3= 5$. The inputs and outputs are either $\left( h, \alpha \right)$ and $t_H$, $\left( h, \alpha \right)$ and $ t_L$, $\left( t_H, t_L \right) $ and $\beta$, or $\left( t_H, t_L \right) $ and $l$.}
\label{NN}
\end{figure}

Once the dataset $(t_H, t_L, h, \alpha, \beta, l)$ is created, we find functions $t_H= f_H (h, \alpha)$, $t_L= f_L (h, \alpha)$, $\beta= f_\beta\left( t_H, t_L \right)$, and $l= f_l\left( t_H, t_L \right)$. Due to the complexity of the behavior of the robot in the above-buckling regime, fitting a simple function to $(t_H, t_L, h, \alpha, \beta, l)$ datapoint is inadequate. Figure~\ref{w_H_traj} shows the trajectory of the robot when $\omega_H$ is applied. Although this complex motion occurs in a length scale significantly smaller than the robot length, since $\beta$ and $\alpha$ critically depend on this trajectory, it is very essential to parameterize it accurately. Here, we employed a three-layer neural network with $N_1= 20$, $N_2= 10$ and $N_3= 5$ neurons for each of these functions and trained the data using Bayesian regularization \cite{Bayes} (see Fig.~\ref{NN}). The number of layers and neurons have been determined by experiment (changing the values of layers and neurons). We did not see significant improvement by using more complex architectures. In particular, we have used an architecture with the same number of layers and twice the number of neurons ($N_1= 40$, $N_2= 20$ and $N_3= 10$), an architecture with four layers ($N_1= 30$, $N_2= 20$, $N_3= 10$, and $N_4= 5$), and an architecture with five layers ($N_1= 50$, $N_2= 30$, $N_3= 20$, $N_4= 10$, and $N_5= 5$). Among the parameters that we have changed during our trainings are: the learning rate, $l_\text{rate} \in \{0.001, 0.01\}$, and the regularization coefficient, $l_\text{reg} \in \{0.001, 0.01, 0.1\}$.

\section{Control}
\label{control}

The goal here is to find the time-dependent angular velocity function $\Omega \left(t_f, {\bf s}(t) \right)$ for time $t_f> t$, given the input ${\bf s}(t)$ at time $t$ (see Eqn.~(\ref{state})). The proposed algorithm in this section assumes the existence of only two control inputs: one below buckling, $\omega_L$, providing a linear trajectory, and one above buckling, $\omega_H$, providing rotation in 3D space. Extension of the single below-buckling angular velocity to multiple (possibly continuously varying) angular velocities is straightforward due to simplicity of the linear trajectory which amounts to only characterization of the linear velocity of the robot along the linear trajectory. The functions $f_H (h, \alpha)$, $f_L (h, \alpha)$, $f_\beta\left( t_H, t_L \right)$, and $f_l\left( t_H, t_L \right)$ used in this section are assumed to be available to the algorithm based on the learning process described in section \ref{inverse}.

If the angular velocity at the current time, $t$, is below buckling ($\omega(t) < \omega_b$), the current trajectory of the robot can be calculated by fitting a line to the head position of the current and $k$ previous time steps, ${\bf x}_0(t- (0:k)\delta t)$, which amounts to solving the following minimization problem
\begin{multline}
\mathcal{L} (a_{1:4})= \min_{a_{1:4}} \sum_{j=0}^k \left[ \left( {\bf x}_{0y}(t-j\delta t)- a_1 {\bf x}_{0x}(t-j\delta t)- a_2 \right)^2 \right. \\ \left. + \left( {\bf x}_{0z}(t-j\delta t)- a_3 {\bf x}_{0x}(t-j\delta t)- a_4 \right)^2 \right] , \label{obj_1} 
\end{multline}
where $a_{1:4}$ are the parameters characterizing the equation of the trajectory; in particular, $y= a_1x+a_2$ and $z= a_3x+a_4$ is the trajectory. Also, ${\bf x}_{0i}$ for $i\in \{x, y, z\}$ is the $i$-th component of ${\bf x}_0$. Equation~(\ref{obj_1}) has the following solutions
\begin{align}
a_1 &= \frac{ (k+1) \sum_{j=0}^k {\bf x}_{0x} (t-j\delta t) {\bf x}_{0y} (t-j\delta t) }{ (k+1) \sum_{j=0}^k {\bf x}_{0x} (t-j\delta t)^2 - \left( \sum_{j=0}^k {\bf x}_{0x} (t-j\delta t) \right)^2 } \nonumber \\ &- \frac{ \sum_{j=0}^k {\bf x}_{0x} (t-j\delta t) \sum_{j=0}^k {\bf x}_{0y} (t-j\delta t) }{ (k+1) \sum_{j=0}^k {\bf x}_{0x} (t-j\delta t)^2 - \left( \sum_{j=0}^k {\bf x}_{0x} (t-j\delta t) \right)^2 } , \nonumber \\ 
a_2 &= \frac{1}{k+1} \left[ \sum_{j=0}^k {\bf x}_{0y} (t-j\delta t) - a_1 \sum_{j=0}^k {\bf x}_{0x} (t-j\delta t) \right] , \nonumber \\
a_3 &= \frac{ (k+1) \sum_{j=0}^k {\bf x}_{0x} (t-j\delta t) {\bf x}_{0z} (t-j\delta t) }{ (k+1) \sum_{j=0}^k {\bf x}_{0x} (t-j\delta t)^2 - \left( \sum_{j=0}^k {\bf x}_{0x} (t-j\delta t) \right)^2 } \nonumber \\ 
&- \frac{ \sum_{j=0}^k {\bf x}_{0x} (t-j\delta t) \sum_{j=0}^k {\bf x}_{0z} (t-j\delta t) }{ (k+1) \sum_{j=0}^k {\bf x}_{0x} (t-j\delta t)^2 - \left( \sum_{j=0}^k {\bf x}_{0x} (t-j\delta t) \right)^2 } , \nonumber \\
a_4 &= \frac{1}{k+1} \left[ \sum_{j=0}^k {\bf x}_{0z} (t-j\delta t) - a_3 \sum_{j=0}^k {\bf x}_{0x} (t-j\delta t) \right] .
\label{a_i}
\end{align}
Equation~(\ref{a_i}) can be used to determine the direction of motion and the degree of linearity, that is, if $ \mathcal{L} (a_{1:4}) > \delta_l$ for a preset small $\delta_l$, then the linear motion assumption is not correct, implying that either the robot is turning (if also $\omega(t) > \omega_b$) or the robot has recently turned but not sufficient time has passed for it to adjust to new below-buckling dynamics (if $\omega(t) < \omega_b$). This information is useful in determining $\Omega \left(t_f, {\bf s}(t) \right)$, since the functions trained in section \ref{inverse} are valid only if the current trajectory is linear. The unit vector determining the direction of motion can be calculated using the result provided in Eqn.~(\ref{a_i}) via
\begin{equation}
{\bf v}= \text{sgn} \{ \left( {\bf x}_0(t)- {\bf x}_1(t) \right) \cdot [1, a_1, a_3] \} \frac{[1, a_1, a_3]}{\big|\big|[1, a_1, a_3]\big|\big|} ,
\label{u}
\end{equation}
where sgn is the sign function.

If the robot follows the successive ${\bf p}_{1:2}$ points exactly, the vector ${\bf v}$ and points ${\bf x}_0 (t)$ and ${\bf p}_{1:2}$ should be coplanar. However, in practice, due to multiple sources of error, noise, or inaccuracy in function training, these points may not be reached exactly. In order to guarantee that the destined points are not compromised due to errors in following the previous trajectories, we define an alternative trajectory at each time, that provided with ${\bf v}$, ${\bf x}_0(t)$, and ${\bf p}_2$ (which define a unique plane together), finds an alternative point ${\bf \hat{p}}_1$ that is as close as possible to ${\bf p}_1$; this amounts to finding the projection of point ${\bf p}_1$ on the plane defined by ${\bf v}$, ${\bf x}_0(t)$, and ${\bf p}_2$ which itself is parameterized via
\begin{equation}
\left[ {\bf v} \times \left( {\bf p}_2- {\bf x}_0(t) \right) \right] \cdot \left( {\bf x}- {\bf x}_0(t) \right)= 0 ,
\label{uqp}
\end{equation}
where ${\bf x}$ is a point in 3D space. The components of the projected point ${\bf \hat{p}}_1$ are
\begin{align}
{\bf \hat{p}}_{1x} &= {\bf p}_{1x}+ \nonumber \\
 & \frac{ \left[ {\bf v} \times \left( {\bf p}_2- {\bf x}_0(t) \right) \right]_x \left[ {\bf v} \times \left( {\bf p}_2- {\bf x}_0(t) \right) \right] \cdot \left( {\bf x}_0(t)- {\bf p}_1 \right) }{ \big|\big| {\bf v} \times \left( {\bf p}_2- {\bf x}_0(t) \right) \big|\big|^2 } , \nonumber \\
{\bf \hat{p}}_{1y} &= \frac{ \left[ {\bf v} \times \left( {\bf p}_2- {\bf x}_0(t) \right) \right]_y }{ \left[ {\bf v} \times \left( {\bf p}_2- {\bf x}_0(t) \right) \right]_x } \left( {\bf \hat{p}}_{1x}- {\bf p}_{1x} \right)+ {\bf p}_{1y} , \nonumber \\
{\bf \hat{p}}_{1z} &= \frac{ \left[ {\bf v} \times \left( {\bf p}_2- {\bf x}_0(t) \right) \right]_z }{ \left[ {\bf v} \times \left( {\bf p}_2- {\bf x}_0(t) \right) \right]_x } \left( {\bf \hat{p}}_{1x}- {\bf p}_{1x} \right)+ {\bf p}_{1z} .
\label{m_hat}
\end{align}
In order to transform the information embedded in the ${\bf x}_0 (t) \rightarrow {\bf \hat{p}}_1 \rightarrow {\bf p}_2$ trajectory into the spherical body-fixed coordinates, we define the following three principal directions (${\bf n}$ here is the same vector defined in Eqn.~(\ref{n_def})):
\begin{equation}
{\bf v}, \ {\bf n}= \frac{{\bf v} \times \left( {\bf x}_1(t)- {\bf x}_2(t) \right)}{\big|\big| {\bf v} \times \left( {\bf x}_1(t)- {\bf x}_2(t) \right) \big|\big|}, \ {\bf w}= \frac{ {\bf v} \times {\bf n}}{\big|\big| {\bf v} \times {\bf n} \big|\big|} .
\label{unv}
\end{equation}
The projection of ${\bf p}_2- {\bf \hat{p}}_1$ vector on these principal directions provides us with the characterization of the trajectory in the body-fixed spherical coordinates. In particular, the four parameters defined below uniquely characterize the immediate desired trajectory:
\begin{align}
h_d &= \big|\big| {\bf p}_2- {\bf \hat{p}}_1 \big|\big| , \nonumber \\
l_d &= \text{sgn} \{ \left( {\bf \hat{p}}_1- {\bf x}_0(t) \right) \cdot {\bf v} \} \big|\big| {\bf \hat{p}}_1- {\bf x}_0(t) \big|\big| , \nonumber \\
\alpha_d &= \frac{180}{\pi}\cos^{-1} \left[ \frac{ \left( {\bf p}_2- {\bf \hat{p}}_1 \right) \cdot {\bf v} }{ \big|\big| {\bf p}_2- {\bf \hat{p}}_1 \big|\big| } \right] , \nonumber \\
\beta_d &= \frac{180}{\pi} \tan^{-1} \left[ \frac{ \left( {\bf p}_2- {\bf \hat{p}}_1 \right) \cdot \left( {\bf v} \times {\bf n} \right) }{ \left( {\bf p}_2- {\bf \hat{p}}_1 \right) \cdot {\bf n} \big|\big| {\bf v} \times {\bf n} \big|\big| } \right] ,
\label{halphabeta}
\end{align}
where the subscript $d$ stands for ``desired" and the angles $\alpha_d$ and $\beta_d$ are represented in degrees units.

Once these parameters are determined, we can find the proper control inputs which amounts to specification of three parameters, $t_\text{app}$, $t_H$, and $t_L$ in the function
\begin{align}
\omega \left(t_f \right) &= \begin{cases}
\omega_L & \text{if } 0 \leq t_f-t < t_\text{app},\\
\omega_H & \text{if }  t_\text{app} \leq t_f-t < t_\text{app}+ t_H, \\ 
\omega_L & \text{if } t_\text{app}+ t_H \leq t_f-t < t_\text{app}+ t_H+ t_L, 
\end{cases}
\label{ctrl_input}
\end{align}
where ``app" corresponds to the time that the steering angular velocity must be applied. For a given $h_d$ and $\alpha_d$, the time intervals $t_H$ and $t_L$ in Eqn.~(\ref{ctrl_input}) can be uniquely determined via
\begin{equation} 
t_H= f_H(h_d, \alpha_d), \ t_L= f_L(h_d, \alpha_d) ,
\label{fs}
\end{equation}
from which we can calculate their corresponding $\beta$ and $l$ via
\begin{equation}
\beta= f_\beta(t_H, t_L), \ l= f_l(t_H, t_L) .
\label{fs_2}
\end{equation}
However, $\beta$, the angle corresponding to $t_H$ and $t_L$, is not necessarily equal to $\beta_d$, and is not controllable either independent of other parameters, once $t_H$ and $t_L$ are determined; the same statement is true for $l$ and $l_d$. The remaining control parameter in Eqn.~(\ref{ctrl_input}), $t_\text{app}$, must be chosen such that $l$ and $\beta$ are as close as possible to $l_d$ and $\beta_d$, respectively. This problem can be formulated as an optimization problem where some $\hat{l}$ and $\hat{\beta}$ that provide a trajectory as close as possible to the desired one are calculated. However, here, by exploiting an important property of the flagellum actuator, its periodic orientation, we only compromise on the value of $l_d$. Note that the orientation of the robot during its linear motion has a period of $60/\omega_L$ seconds ($\omega_L$ is in rpm units), that is
\begin{equation}
{\bf n} (t+ 60/\omega_L)= {\bf n} (t) .
\end{equation}
On the other hand, the robot travels a distance of $60v_{\omega_L}/\omega_L$ during one period, where $v_{\omega_L}$ is the linear velocity when $\omega_L$ is applied. This is a very small distance compared to the size of the robot; for instance, for the parameters considered in this work, this distance is about 4 mm while the length of the flagellum is 200 mm. Taking into account these properties of the system, we choose the following $t_\text{app}$
\begin{equation}
t_\text{app}= \frac{\beta_d- \beta}{6\omega_L}+ \frac{60}{\omega_L} \text{arg}\min_\kappa \left\{ \Big| \frac{\beta_d- \beta+ 360\kappa}{6\omega_L} v_{\omega_L}+ l- l_d \Big| \right\} .
\label{t_app_1}
\end{equation}
The first term in Eqn.~(\ref{t_app_1}) guarantees that the robot will be able to follow $\beta_d$ accurately, while the second term (the $\text{arg}\min$ expression) guarantees that $l_d$ is reproduced as close as possible. The solution to Eqn.~(\ref{t_app_1}) is 
\begin{equation}
t_\text{app}= \frac{\beta_d- \beta}{6\omega_L}+ \frac{60}{\omega_L} \text{round} \left\{ \frac{\omega_L (l_d- l)}{60v_{\omega_L}}- \frac{\beta_d- \beta}{360} \right\} ,
\label{t_app_2}
\end{equation}
where ``round" implies rounding to the nearest integer. This approximation leads to the maximum error of $30v_{\omega_L}/\omega_L$ in following $l_d$ which is very negligible; for instance, it is 2 mm (compare with the length of the flagellum which is 200 mm) at $\omega_L= 3$ rpm (the value chosen in our work).

The control algorithm $\Omega(t_f, {\bf s}(t))$ based on the results of this section is provided in Algorithm \ref{Algo}. The time used in this algorithm, $t$, is the universal clock, therefore, after each time step, we have $t \gets t+\delta t$. The algorithm first initializes the angular velocity of the first $k$ time steps to $\omega_L$ to allow sufficient time from the initial non-linearity associated with starting from the static state. Then it checks whether all the elements of ${\bf s}(t)$ are non-empty, and if this is true, it proceeds to check whether the assumption of linear trajectory is valid, since otherwise, the results provided in this section and section \ref{inverse} do not hold anymore. If the algorithm passes all these tests, it calculates the future angular velocities and stores them in $\omega(t_f)$. Note that here we are assuming that the algorithm does not erase any of its information over the sequence of time steps. For instance, if there is incoming information of ${\bf p}_{1:2}$ while the robot is turning (${\cal L}(a_{1:4}) > \delta_l$), it stores ${\bf p}_{1:2}$ and will use it as soon as other conditions hold as well. The same is true for the parameter $j$ in the algorithm. Finally, if the condition $\omega(t+\delta t)= \emptyset$ is satisfied at time $t$, it sets $\omega(t+\delta t)= 0$ to stop the robot ($\emptyset$ is the empty set).

\begin{algorithm}[t]
\caption{Online control $\Omega(t_f, {\bf s}(t))$}\label{Algo}
\begin{algorithmic}
\REQUIRE {${\bf s} (t)$: input (see Eqn.~(\ref{state}))}
\IF{$t=0$}
\FOR{$j=1$ to $j=k$}
\STATE $\omega(t+j\delta t)= \omega_L$
\ENDFOR
\ENDIF
\IF{$t>k\delta t$, ${\bf x}_1(t) \neq \emptyset$, ${\bf x}_2(t) \neq \emptyset$, ${\bf p}_1 \neq \emptyset$, ${\bf p}_2 \neq \emptyset$, and $\forall i \in \{0,...,k\}: {\bf x}_0 (t- i\delta t) \neq \emptyset$}
\STATE Calculate $a_{1:4}$ from Eqn.~(\ref{a_i}); calculate ${\cal L}(a_{1:4})$ from Eqn.~(\ref{obj_1});
\IF{${\cal L}(a_{1:4}) \leq \delta_l$}
\STATE Calculate ${\bf v}$ from Eqn.~(\ref{u}); ${\bf \hat{p}}_1$ from Eqn.~(\ref{m_hat}); $h_d$, $l_d$, $\alpha_d$, and $\beta_d$ from Eqn.~(\ref{halphabeta}); $t_H$ and $t_L$ from Eqn.~(\ref{fs}); $\beta$ and $l$ from Eqn.~(\ref{fs_2}); $t_\text{app}$ from Eqn.~(\ref{t_app_2})
\STATE $j \gets 1$
\WHILE{$j\delta t < t_\text{app}+ t_H+ t_L$}
\IF{$j\delta t \geq t_\text{app}$ and $j\delta t < t_\text{app}+ t_H$}
\STATE $\omega(t+j\delta t) \gets \omega_H$
\ELSE
\STATE $\omega(t+j\delta t) \gets \omega_L$
\ENDIF
\STATE $j \gets j+1$
\ENDWHILE
\ENDIF
\ENDIF
\STATE $\omega(t_f) \gets \omega(t+(1:j)\delta t)$
\STATE $j \gets j-1$
\IF{$\omega(t_f)= \emptyset$}
\STATE $\omega(t_f)=0$
\ENDIF
\RETURN {$\omega(t_f)$}
\end{algorithmic}
\end{algorithm}

\section{Results}
\label{result}

\begin{figure}[t]
\centerline{\psfig{figure=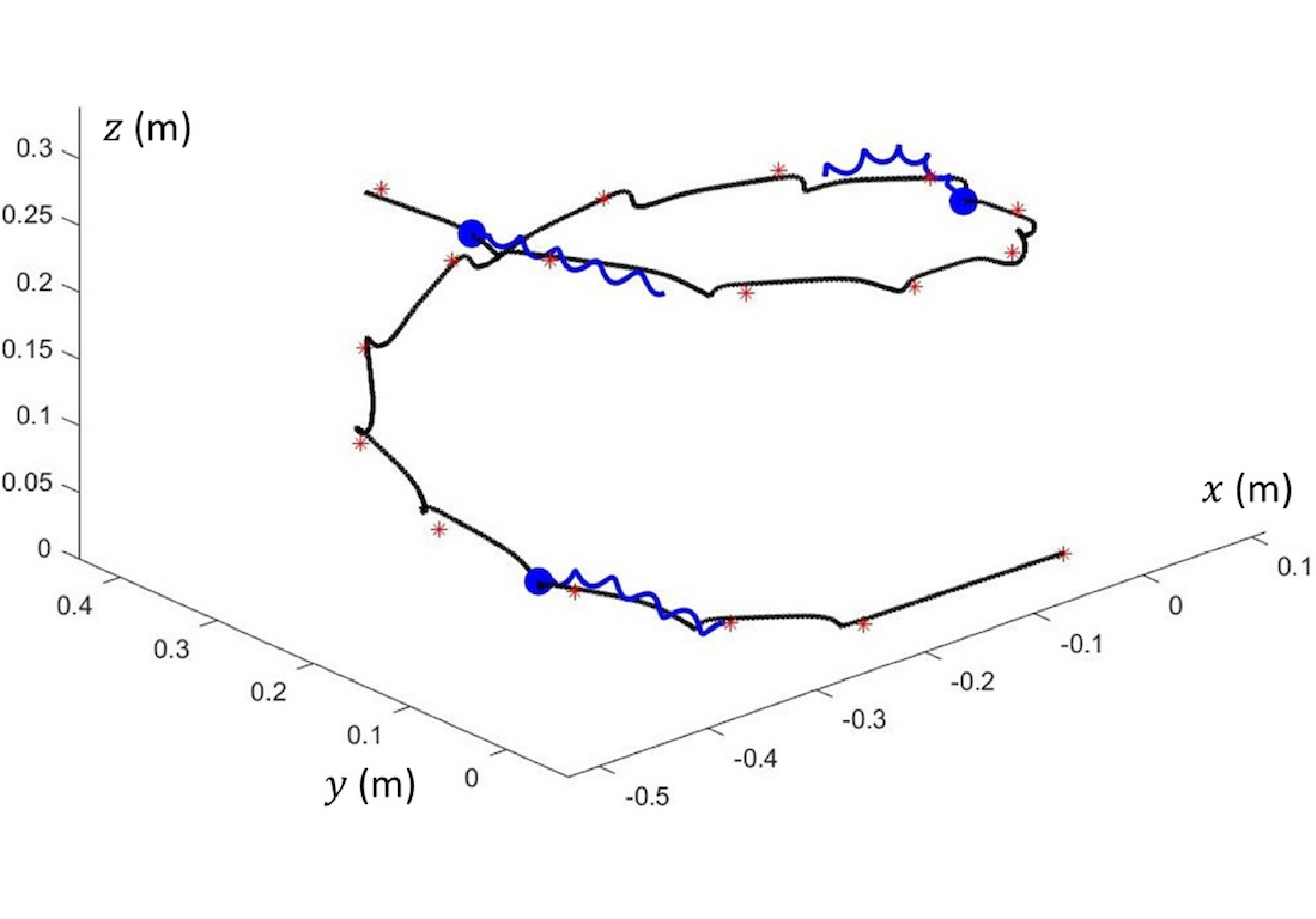,width=3.0in}}
\caption{ The robot trajectory (the helical path) and the points that it was expected to follow (the asterisks).}
\label{path_H_robo}
\end{figure}

We have used Algorithm \ref{Algo} to follow a helical path. The points characterizing the desired trajectory (the sequence ${\bf p}_{1:2}$), the robot trajectory, and a few screenshots of the robot itself are shown in Fig.~\ref{path_H_robo}. An animation of this simulation can also be found in this \href{https://youtu.be/WbI9zL-ogM8}{\color{red}{link}}. We observe that overall the robot has been able to follow the desired points with reasonable accuracy. The robot head (the spheres in the figure) is the point that is expected to follow the desired points (the asterisks) sequentially.

\begin{figure}[t]
\centerline{\psfig{figure=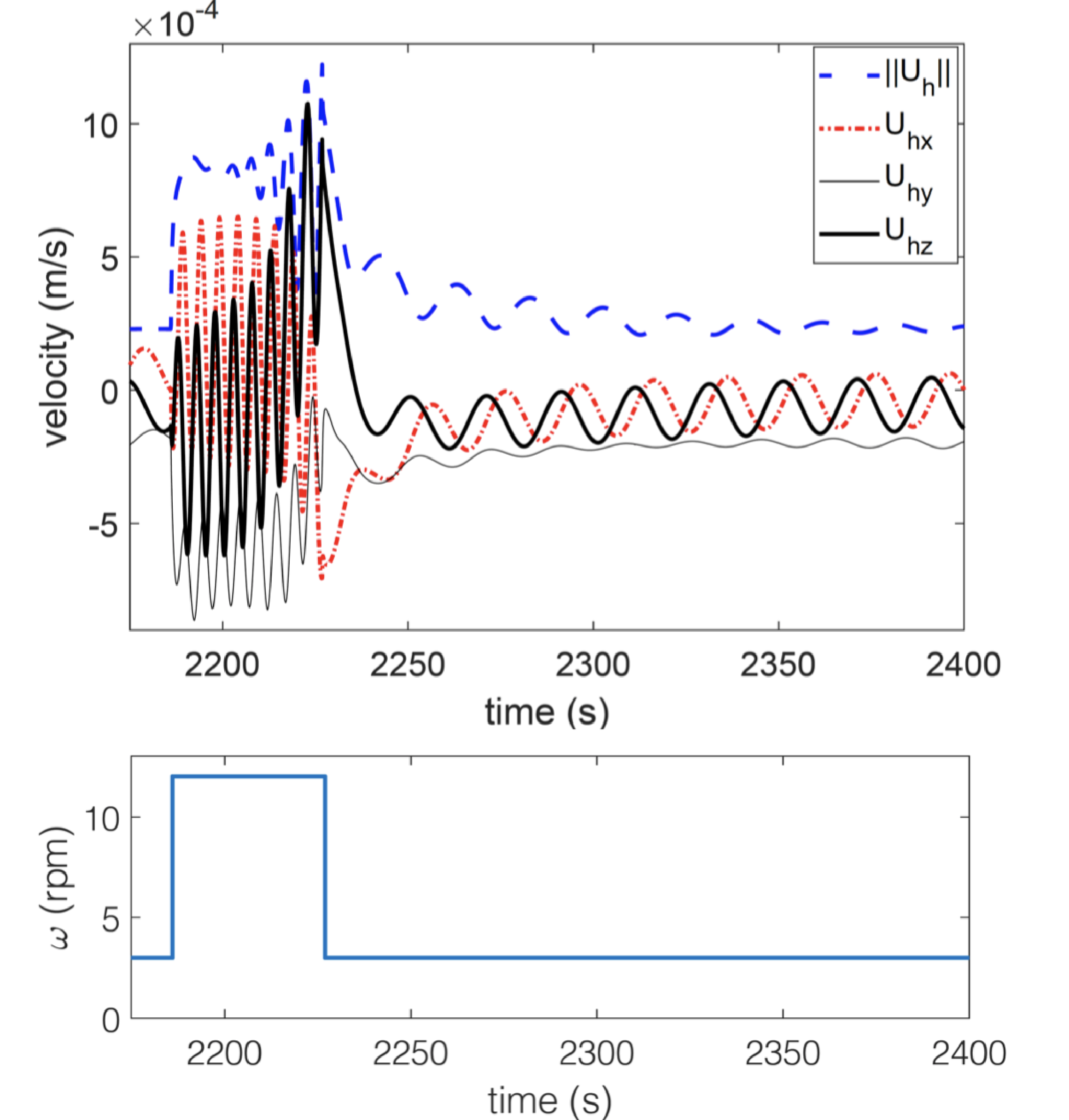,width=3.2in}}
\caption{ Top: output (velocity of the head and its components), and bottom: input (angular velocity) during a steering incident. }
\label{In-out}
\end{figure}

Similar to the time duration required for applying the above-buckling angular velocity for providing a given steering angle, the time taken for the flagellum to reach its unbent straight-line configuration also varies depending on the material property of the flagellum. Figure~\ref{In-out}, top, shows the robot velocity magnitude and its three components during a steering incident and Fig.~\ref{In-out}, bottom, shows its corresponding input angular velocity. The complexity of the robot path when the above-buckling angular velocity is applied is well captured in Fig. \ref{In-out} (top). Another important observation is the long post-buckling effect in robot trajectory and velocity which lasts about three times the applied $\omega_H$ duration. This effect is the reason that in section \ref{inverse} we have allowed at least 100 s to pass from $t_0$ in order to extract the data from the generated (long) trajectories. This period can be shortened by performing an optimization over different $\omega_L$ and $\omega_H$ values, different material properties, and different structural properties, aimed at minimizing the total buckling effect duration. A similar inverse problem framework can be utilized to identify the relationship between these inputs and the robot trajectory.

\begin{figure*}[t]
\centerline{\psfig{figure=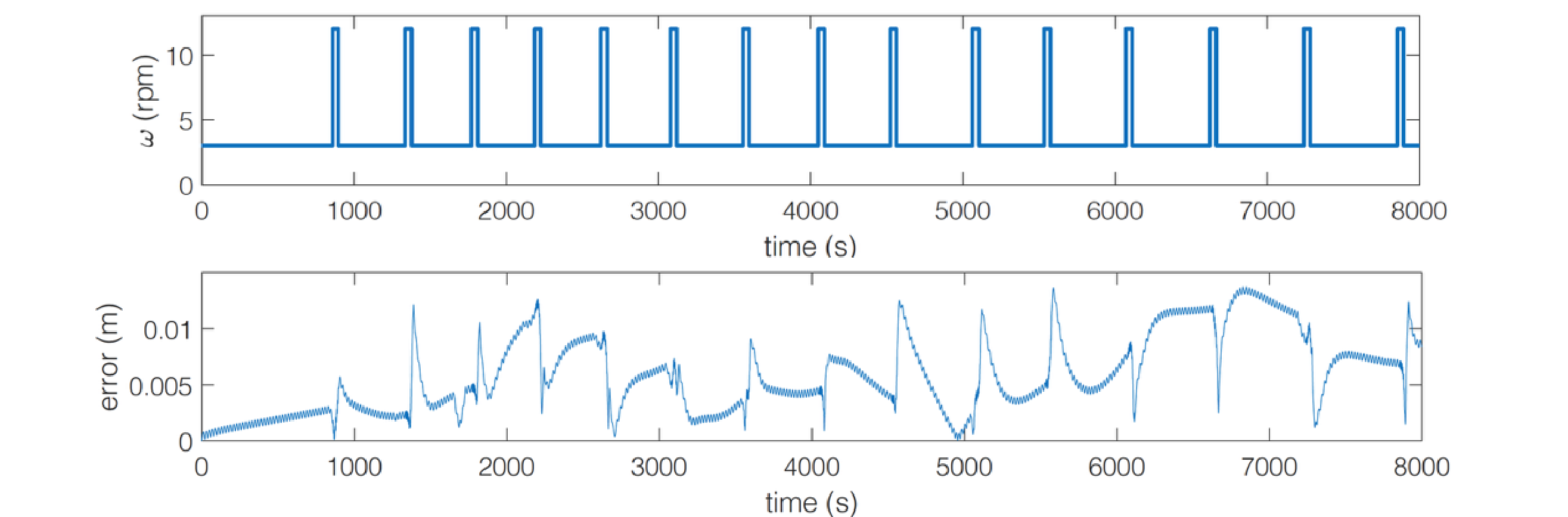,width=5.8in}}
\caption{Control input (top), and the tracking error (bottom) corresponding to the trajectory of Fig.~\ref{path_H_robo}.}
\label{long_time}
\end{figure*}

One of the advantages of the algorithm used here is that it has binary input, $\omega_L$ and $\omega_H$, which simplifies both control policy and design of the actuator. Figure~\ref{long_time}, top, shows the control input used for generation of the trajectory of Fig.~\ref{path_H_robo}. We observe that the periods at which the above-buckling angular velocity has been input are significantly shorter than the below-buckling periods. This is also partly due to small steering angles, $\alpha$, used for the trajectory of Fig.~\ref{path_H_robo}. Larger values of $\alpha$ require longer periods of $\omega_H$. These periods also depend on the material and structural properties of the robot and the fluid in which it is operating. We have also plotted the tracking error in Fig.~\ref{long_time}, bottom. The tracking error in this figure is defined as the minimum distance (Euclidean distance) between robot head and the prescribed helical path. We observe that overall the error is small. We also notice that for the most part, the largest errors correspond to steering incidents---the same trend observed in Fig.~\ref{sketch_2}. This error can decrease if a set of material properties is chosen such that the deviation of robot trajectory from coplanar before- and after- steering lines is minimal. The overall error may also be reduced by generating more datapoints during the learning process, choosing a smaller linearity criteria, $\delta_l$, or increasing the time window for estimation of direction of motion (increasing $k$).

\section{Conclusion}
\label{conclusion}

Inspired by motion of bacteria, we have proposed a method for locomotion of robots in low Reynolds number using one single input. Restricting the angular velocity to obtain a binary input ($\omega_L$ and $\omega_H$) provides simple design of actuators, addressing one of the main challenges that millimeter/sub-millimeter robots face. By combining an accurate numerical model of the dynamics of the helical filament with neural network, we have been able to capture the complexity associated with the inverse dynamics, bypassing the issues associated with approximate analytical/semi-analytical approaches to characterization of the robot motion. Since it has been verified previously that the simulation tools employed here resemble the dynamics of the helical filament accurately, in the present work, we have primarily focused on the numerical and algorithmic aspects of the motion control. Our future emphasis is on the implementation of the algorithm provided here on small scale soft robots.

In this work, we have focused on uniflagellar robots due to the simplicity of their input (binary input) which will consequently lead to simpler design of actuators. Such designs are suitable when the primary objective of the robot is reaching a target, such as imaging or sensing. However, higher maneuverability of robot requires more complicated inputs. Here, we did not assess the instability in motion due, for instance, to carrying a cargo, such as drug delivery tasks. Such designs require both more inputs, such as multiple flagella instead of one, and more advanced control policy such as adaptive and/ or sliding mode control strategies \cite{slotine}. In future, we will study both algorithmic and experimental aspects of such designs.

In the present work, we proposed an algorithm assuming perfect knowledge of the system state which consequently led to designs based on only the next two points of the desired trajectory (assuming that the state of the system is readily available in future time steps). If the system state is imperfect, for instance due to communication and sensory issues, the algorithm may need to output angular velocity function for more than one turning event (trajectory characterization with more than two points). Model predictive control strategies (MPC) based on online sequential optimization of the given sequential trajectories can address such issues \cite{MPC}; however, here, due to computational cost of the forward dynamics simulation, we did not study them. In future, we will explore further aspects of the numerical method used to solve the forward dynamics in order to reduce its cost and consequently use MPC for sequential motion control.

\section*{Acknowledgment}
We acknowledge support from the Henry Samueli School of Engineering and Applied Science, University of California, Los Angeles.

%
% ...

\bibliographystyle{asmems4}

\bibliography{asme2e}

\begin{thebibliography}{1}

\bibitem{Mukherjee_1} { A. Hellum, R. Mukherjee, A. Benard, and A.J. Hull}, {\em Modeling and simulation of the dynamics of a submersible propelled by a fluttering fluid-conveying tail}, Journal of Fluids and Structures 36, 83-110 (2013).

\bibitem{Mukherjee_2} { P.C. Strefling, A.M. Hellum, and R. Mukherjee}, {\em Modeling, simulation, and performance of a synergistically propelled ichthyoid}, IEEE/ASME Transactions on Mechatronics 17, 36-45 (2012).

\bibitem{app_Ornes} { S. Ornes}, {\em Medical microrobots have potential in surgery, therapy, imaging, and diagnostics}, Proceedings of the National Academy of Sciences 114, 12356-12358 (2017).

\bibitem{surgery_Taylor} { R. H. Taylor, D. Stoianovici}, {\em Medical robotics in computer-integrated surgery}, IEEE Transactions on Robotics and Automation 19, 765-781 (2003).

\bibitem{prop_small} { J. J. Abbott, Z. Nagy, F. Beyeler, and B. J. Nelson}, {\em Robotics in the Small}, IEEE Robotics and Automation Magazine 14, 92-103 (2007).

\bibitem{prop_Nain} { S. Nain and N. N. Sharma}, {\em Propulsion of an artificial nanoswimmer: a comprehensive review}, Frontiers in Life Science 8, 2-17 (2015).

\bibitem{drug_Nelson} { B. J. Nelson, I. K. Kaliakatsos, and J. J. Abbott}, {\em Microrobots for minimally invasive medicine}, Annual Review of Biomedical Engineering 12, 55-85 (2010).
 
\bibitem{drug_Li} { H. Li, J. Tan, and M. Zhang}, {\em Dynamics modeling and analysis of a swimming microrobot for controlled drug delivery}, IEEE Transactions on Robotics and Automation 6, 220-227 (2009).

\bibitem{drug_Fusco} { S. Fusco, F. Ullrich, J. Pokki, G. Chatzipirpiridis, B. Ozkale, K. M. Sivaraman, O. Ergeneman, S. Pane, and B. J. Nelson}, {\em Microrobots: a new era in ocular drug delivery.}, Expert Opinion on Drug Delivery 11, 1815-1826 (2014).

\bibitem{cargo_Medina} { M. Medina-Sanchez, L. Schwarz, A. K. Meyer, F. Hebenstreit, and O. G. Schmidt}, {\em Cellular cargo delivery: Toward assisted fertilization by sperm-carrying micromotors}, Nano Letters 16, 555-561 (2016).

\bibitem{surgery_Edd} { J. Edd, S. Payen, B. Rubinsky, M. L. Stoller, and M. Sitti}, {\em Biomimetic propulsion for a swimming surgical micro-robot}, IEEE International Conference on Intelligent Robots and Systems, Las Vegas, Nevada (2003).

\bibitem{image_Nelson} { A. Servant, F. Qiu, M. Mazza, K. Kostarelos, and B. J. Nelson}, {\em Controlled in vivo swimming of a swarm of bacteria-Like microrobotic flagella}, Advanced Materials 27, 2981-2988 (2015).

\bibitem{prop_Pak} { O. S. Pak, W. Gao, J. Wang, and E. Lauga}, {\em High-speed propulsion of flexible nanowire motors: Theory and experiments}, Soft Matter 7, 8169-8181 (2011).

\bibitem{prop_Ghosh} { A. Ghosh and P. Fischer}, {\em Controlled propulsion of artificial magnetic nanostructured propellers}, Nano Letters 9, 2243-2245 (2009).

\bibitem{prop_Abbot} { J. J. Abbott, K. E. Peyer, M. C. Lagomarsino, L. Zhang, L. Dong, I. K. Kaliakatsos, and B. J. Nelson}, {\em How should microrobots swim?}, The International Journal of Robotics Research 28, 1434-1447 (2009).

\bibitem{prop_Magdanz} { V. Magdanz, S, Sanchez, and O. G. Schmidt}, {\em Development of a sperm-flagella driven micro-bio-robot}, Advanced Materials 25, 6581-6588 (2013).

\bibitem{prop_Bell} { D. J. Bell, S. Leutenegger, K. M. Hammar, L. X. Dong, and B. J. Nelson}, {\em Flagella-like propulsion for microrobots using a nanocoil and a rotating electromagnetic field}, IEEE International Conference on Robotics and Automation, Roma, Italy (2007).

\bibitem{ctrl_Nour} { H. Nourmohammadi and J. Keighobadi}, {\em Design, modeling and control of a maneuverable swimming micro-robot}, Proceedings of the 19th International Federation of Automatic Control  World Congress, Cape Town, South Africa (2014).

\bibitem{ctrl_Lobaton} { E. J. Lobaton and A. M. Bayen}, {\em Modeling and optimization analysis of a single-flagellum micro-structure
through the method of regularized Stokeslets}, IEEE Transactions on Control Systems Technology 17, 907-916 (2009).

\bibitem{wave_Behkam} { B. Behkam and M. Sitti}, {\em Design methodology for biomimetic propulsion of miniature swimming robots}, Journal of Dynamic Systems, Measurement, and Control 128, 36-43 (2005).

\bibitem{distributed} {G. Kosa, M. Shoham, and M. Zaaroor}, {\em Propulsion method for swimming microrobots}, IEEE Transactions on Robotics 23, 137-150 (2007).

\bibitem{chem} {S. Sanchez, L. Soler, and J. Katuri}, {\em Chemically powered Micro- and Nanomotors}, Angewandte Chemie International Edition 54, 1414-1444 (2015).

\bibitem{magnet_Kim} { S. Kim, S. Lee, J. Lee, B. J. Nelson, L. Zhang, and H. Choi}, {\em Fabrication and manipulation of ciliary microrobots with non-reciprocal magnetic actuation}, Scientific Reports 6, 30713 (2016).

\bibitem{2D} {R. W. Carlsen, M. R. Edwards, J. Zhuang, C. Pacoret,
and M. Sitti}, {\em Magnetic steering control of multi-cellular
bio-hybrid microswimmers}, Lab on a Chip 14, 3850-3859 (2014).

\bibitem{electric_Osada} { Y. Osada, H. Okuzaki, and H. Hori}, {\em A polymer gel with electrically driven motility}, Nature 355, 242-244 (1992).

\bibitem{RBC} {Z. Wu, B. Esteban-Fernández de Avila, A. Martin, C. Christianson, W. Gao, S. K. Thamphiwatana, A. Escarpa, Q. He, L. Zhang, and J. Wang}, {\em RBC micromotors carrying multiple cargos towards potential theranostic applications}, Nanoscale 7, 13680-13686 (2015).

\bibitem{multi_Nguyen} { F. T. M. Nguyen and M. D. Graham}, {\em Impacts of multiﬂagellarity on stability and speed of bacterial locomotion}, 
arXiv:1805.00081 (2018).

\bibitem{poly_Ali} { J. Ali, U K. Cheang, J. D. Martindale, M. Jabbarzadeh, H. C. Fu, and M. Jun Kim}, {\em Bacteria-inspired nanorobots with flagellar polymorphic transformations and bundling}, Scientific Reports 7, 14098 (2017).

\bibitem{bundle_Hintsche} { M. Hintsche, V. Waljor, R. Grobmann, M. J. Kuhn, K. M. Thormann, F. Peruani, and C. Beta}, {\em A polar bundle of flagella can drive bacterial swimming by pushing, pulling, or coiling around the cell body}, Scientific Reports 7, 16771 (2017).

\bibitem{buckle_Nguyen} { F. T. M. Nguyen and M. D. Graham}, {\em Buckling instabilities and complex trajectories in a simple model of uniflagellar bacteria}, Biophysical Journal 112, 1010-1022 (2017).

\bibitem{buckle_Son} { K. Son, J. S. Guasto, and R. Stocker}, {\em Bacteria can exploit a flagellar buckling instability to change direction}, Nature Physics 9, 494-498 (2013).

\bibitem{rodenborn2013propulsion} {B. Rodenborn, C.-H. Chen, H. L. Swinney, B. Liu, and H. P. Zhang}, {\em Propulsion of microorganisms by a helical flagellum}, Proceedings of the National Academy of Sciences 110, E338--E347 (2013).

\bibitem{kim2005deformation} { M. Kim and T. R. Powers}, {\em Deformation of a helical filament by flow and electric or magnetic fields}, Physical Review E 71, 021914 (2005).

\bibitem{bergou2008discrete} {	M. Bergou, M. Wardetzky, S. Robinson, B. Audoly, and	E. Grinspun}, {\em Discrete elastic rods}, ACM Transactions on Graphics (TOG) 27, 63 (2008).

\bibitem{bergou2010discrete} { M. Bergou, B. Audoly,	 E. Vouga, M. Wardetzky, and E. Grinspun}, {\em Discrete viscous threads}, ACM Transactions on Graphics (TOG) 29, 116 (2010).

\bibitem{jawed2018primer} {M. K. Jawed, A. Novelia, and O. M. O\rq{}Reilly}, {\em A primer on the kinematics of discrete elastic rods}, Springer (2018).

\bibitem{lighthill1976flagellar} {J. Lighthill}, {\em Flagellar hydrodynamics}, SIAM Review 18, 161-230 (1976).

\bibitem{jawed2015propulsion} {M. K. Jawed, N. K. Khouri, F. Da, E. Grinspun, and P. M. Reis}, {\em Propulsion and instability of a flexible helical rod rotating in a viscous fluid}, Physical Review Letters 115 (16), 168101 (2015).

\bibitem{jawed2017dynamics} {M. K. Jawed and P. M. Reis}, {\em Dynamics of a flexible helical filament rotating in a viscous fluid near a rigid boundary}, Physical Review Fluids 2, 034101 (2017).

\bibitem{higdon1979hydrodynamic} {J. J. L. Higdon}, {\em A hydrodynamic analysis of flagellar propulsion}, Journal of Fluid Mechanics 90, 685-711 (1979).

\bibitem{thawani2018trajectory} {A. Thawani and M. S. Tirumkudulu}, {\em Trajectory of a model bacterium}, Journal of Fluid Mechanics 835, 252-270 (2018).

\bibitem{huang2018numerical} {W. Huang and M. K. Jawed}, {\em Numerical exploration on buckling instability for directional control in flagellar propulsion}, Accepted, Soft Matter, DOI: 10.1039/C9SM01843C (2019).

\bibitem{robot_inverse} { S-Y. Kung and J. N. Hwang}, {\em Neural network architectures for robotic applications}, IEEE Transactions on Robotics and Automation 5, 641-657 (1989).

\bibitem{source_inverse} { R. Khodayi-mehr, W. Aquino, and M. M. Zavlanos}, {\em Model-based active source identification in complex environments}, arXiv:1706.01603 (2018). 

\bibitem{driver_inverse} { M. Forghani, J. M. McNew, D. Hoehener, and D. Del Vecchio}, {\em Safety control of a class of stochastic order preserving systems with application to collision avoidance near stop signs}, American Control Conference, Chicago, IL (2015).

\bibitem{human_inverse} { E. De Momi, L. Kranendonk, M. Valenti, N. Enayati, and G. Ferrigno}, {\em A neural network-based approach for trajectory planning in robot–human handover tasks}, Frontiers in Robotics and AI 3, 34 (2016).  

\bibitem{reaction_inverse} { M. Forghani, J. M. McNew, D. Hoehener, and D. Del Vecchio}, {\em Design of driver-assist systems under probabilistic safety specifications near stop signs}, IEEE Transactions on Automation Science and Engineering 13, 43-53 (2016).

\bibitem{price_pred} {P. Rezazadeh Kalehbasti, L. Nikolenko, and H. Rezaei}, {\em Airbnb Price Prediction Using Machine Learning and Sentiment Analysis}, arXiv:2010.01611 (2019).

\bibitem{kirchhoff1859uber} {G. Kirchhoff}, {\em Ueber das Gleichgewicht und die Bewegung eines unendlich d\"{u}nnen elastischen Stabes}, Journal f\"{u}r die Reine und Angewandte Mathetmatik 56, 285-313 (1859).

\bibitem{vogel2012motor} {R. Vogel and H. Stark}, {\em Motor-driven bacterial flagella and buckling instabilities}, European Physical Journal E 35, 15 (2012).

\bibitem{Bayes} { F. D. Foresee and M. T. Hagan}, {\em Gauss-Newton approximation to Bayesian learning}, Proceedings of the International Conference on Neural Networks, Houston, TX (1997).

\bibitem{slotine} {J.-J. E. Slotine and W. Li}, {\em Applied nonlinear control}, Englewood Cliffs, New Jersey: Prentice-Hall (1991). 

\bibitem{MPC} {M. Morari and J. H. Lee}, {\em Model predictive control: past, present and future}, Computers and Chemical Engineering 23, 667-682 (1999). 

\end{thebibliography}

\appendix       
\section*{Appendix A: Parameterization of the Generated Trajectories}
\label{paramet}
Here, we provide the details of converting the information embedded in the generated trajectories, ${\bf x}_0(t_0 \leq t< t_0+ t_H+ t_L)$ into the form of the data we use in the control algorithm, $(t_H, t_L, h, \alpha, \beta, l)$. Since the before-turning trajectory, initial direction of motion, and initial orientation of the robot do not depend on its steering angles and consequently inputs $t_H$ and $t_L$, Eqns.~(\ref{obj_1})--(\ref{u}) and Eqn.~(\ref{unv}) are valid here. Therefore, $a_{1:4}$ can be calculated from Eqn.~(\ref{a_i}) with $t$ replaced by $t_0$, and in the same way, ${\bf v}$, ${\bf n}$, and ${\bf w}$ can be calculated from Eqns.~(\ref{u}) and (\ref{unv}) with $t$ replaced by $t_0$.

The next step is to parameterize the trajectory of the robot after the turning has stopped, that is, finding the unknowns in the linear relationships $y=b_1x+b_2$ and $z=b_3x+b_4$, which parameterize the after-turning linear trajectory. This amounts to satisfying the following conditions 
\begin{align}
\begin{cases}
a_1x+ a_2= b_1x+ b_2 \ \text{and}\ a_3x+ a_4= b_3x+ b_4,\\
b_1{\bf x}_{0x} (t_e)+ b_2= {\bf x}_{0y} (t_e) \ \text{and}\ b_3{\bf x}_{0x} (t_e)+ b_4= {\bf x}_{0z} (t_e),
\end{cases}
\label{b_eq}
\end{align}
where the notation $t_e= t_0+ t_H+ t_L$ (the end point) has been used for compactness. The first condition in Eqn.~(\ref{b_eq}) guarantees that the trajectories before and after turning are coplanar, leading to one equation (after setting the $x$ in two equations equal to each other), and the second conditions guarantee that the end position is on the trajectory, leading to two additional equations; in fact, ${\bf x}_0 (t_e)$ for different values of $t_e$ corresponds to different ${\bf p}_2$ points in Fig.~\ref{sketch_2}. The fourth equation required to determine $b_{1:4}$ is obtained by fitting the best line in the mentioned form ($y=b_1x+b_2$ and $z=b_3x+b_4$) to the trajectory of the robot, that is, solving the following optimization problem
\begin{multline}
\min_{b_{1:4}} \sum_{j=1}^m \left[ \left( {\bf x}_{0y}(t_e -j\delta t)- b_1 {\bf x}_{0x}(t_e -j\delta t)- b_2 \right)^2 \right. \\ \left. + \left( {\bf x}_{0z}(t_e -j\delta t)- b_3 {\bf x}_{0x}(t_e -j\delta t)- b_4 \right)^2 \right] , \label{obj_2} 
\end{multline}
where $m= \text{round} \left\{ \frac{t_e- k\delta t}{\delta t} \right\}$. Solving Eqns.~(\ref{b_eq}) and (\ref{obj_2}) leads to following solutions
\begin{align}
b_1 &= \nonumber \\ &\frac{ \sum_{j=1}^m \left[ {\bf x}_{0x}(t_e-j\delta t)- {\bf x}_{0x}(t_e) \right] \left[ {\bf x}_{0y}(t_e -j\delta t)- {\bf x}_{0y}(t_e) \right] }{1+ \chi^{-2}} \nonumber \\ & +\frac{ \left( a_1-a_3\chi \right) \sum_{j=1}^m \left[ {\bf x}_{0x}(t_e- j\delta t)- {\bf x}_{0x}(t_e) \right]^2 }{1+ \chi^2}+ \nonumber \\ & \frac{ \sum_{j=1}^m \left[ {\bf x}_{0x}(t_e- j\delta t)- {\bf x}_{0x}(t_e) \right] \left[ {\bf x}_{0z}(t_e- j\delta t)- {\bf x}_{0z}(t_e)  \right] }{\chi+ \chi^{-1}} , \nonumber \\
b_2 &= {\bf x}_{0y} (t_e)- b_1{\bf x}_{0x} (t_e) , \nonumber \\
b_3 &= \frac{\left( a_1- b_1 \right) \left( {\bf x}_{0z}(t_e)- a_4 \right) - a_3 \left( b_2- a_2 \right)}{ \left( a_1- b_1 \right) {\bf x}_{0x}(t_e)- b_2+ a_2} , \nonumber \\ 
b_4 &= {\bf x}_{0z} (t_e)- b_3{\bf x}_{0x} (t_e) ,
\label{b_i}
\end{align}
where 
\begin{equation}
\chi= \frac{a_1{\bf x}_{0x} (t_e)+ a_2- {\bf x}_{0y} (t_e)}{ a_3{\bf x}_{0x} (t_e)+ a_4- {\bf x}_{0z} (t_e)} .
\label{chi}
\end{equation}
Next, we find the location of the intersection of the linear trajectories before and after the rotation and the projection of ${\bf x}_0 (t_0)$ on the $y= a_1x+a_2$ and $z=a_3x+a_4$ lines\footnote{Note that although the point ${\bf x}_0 (t_0)$ is expected to lie on $y= a_1x+a_2$ and $z=a_3x+a_4$ lines, as it was explained previously, due to helical path of the robot in the linear regime (Fig.~\ref{sketch_2}), ${\bf x}_0 (t_0)$ is slightly distorted from its image on the lines.}. The coordinates of this projected point can be calculated using the following  
\begin{align}
\begin{cases}
{\bf \hat{x}}_{0x}= \frac{{\bf x}_{0x} (t_0)+ a_1\left( {\bf x}_{0y} (t_0)- a_2 \right)+ a_3 \left( {\bf x}_{0z} (t_0)- a_4 \right)}{1+ a_1^2+ a_3^2} ,\\
{\bf \hat{x}}_{0y}= a_1{\bf \hat{x}}_{0x}+ a_2 , \\ {\bf \hat{x}}_{0z}= a_3{\bf \hat{x}}_{0x}+ a_4 ,
\end{cases}
\label{x1y1z1}
\end{align}
from which the position of the intersection can also be calculated in the following form
\begin{equation}
{\bf p}_1= \left[ \frac{b_2- a_2}{a_1- b_1} , \frac{a_1b_2- b_1a_2}{a_1- b_1},  \frac{a_3b_4- b_3a_4}{a_3- b_3} \right] .
\label{x2y2z2}
\end{equation}
The parameters $h$, $\alpha$, $\beta$, and $l$ used in Eqns.~(\ref{fs}) and (\ref{fs_2}) can be calculated based on equations similar to those used in Eqn.~(\ref{halphabeta})\footnote{The parameters $h$ and $\alpha$ used here are similar to $h_d$, $\alpha_d$ used in Eqn.~(\ref{fs}), but in order to distinguish between the desired parameters input to the control algorithm and the ones obtained based on the data training, we have used subscript $d$}. Also, we can replace ${\bf x}_0(t_e)$ with ${\bf p}_2$, therefore,
\begin{align}
h &= \big|\big| {\bf p}_2 - {\bf p}_1 \big|\big| , \nonumber \\
l &= \text{sgn} \{ \left( {\bf p}_1- {\bf \hat{x}}_0 \right) \cdot {\bf v} \} \big|\big| {\bf p}_1- {\bf \hat{x}}_0 \big|\big| , \nonumber \\
\alpha &= \frac{180}{\pi}\cos^{-1} \left[ \frac{ \left( {\bf p}_2- {\bf p}_1 \right) \cdot {\bf v} }{ \big|\big| {\bf p}_2- {\bf p}_1 \big|\big| } \right] , \nonumber \\
\beta &= \frac{180}{\pi} \tan^{-1} \left[ \frac{ \left( {\bf p}_2- {\bf p}_1 \right) \cdot \left( {\bf v} \times {\bf n} \right) }{ \left( {\bf p}_2- {\bf p}_1 \right) \cdot {\bf n} \big|\big| {\bf v} \times {\bf n} \big|\big| } \right] 
\label{halphabetadell}
\end{align}
The set $(t_H, t_L, h, \alpha, \beta, l)$ is one unique datapoint used in the neural network training algorithm.

An alternative to training two separate functions, $t_H= f_H (h, \alpha)$ and $t_L= f_L (h, \alpha)$, is to train one function $\left( t_H, t_L \right)= f_{H,L} (h, \alpha)$. An important consideration in identification of function $\left( t_H, t_L \right)= f_{H,L} (h, \alpha)$ is to ensure that the trained network is not biased toward either of the outputs. A reasonable characterization of the error between the true functional relationship of the datapoint inputted to the network and the trained function is the error observed in the value of the final target point ${\bf p}_2$ (see Fig.~\ref{sketch_2}), since the position of ${\bf p}_2$ is much more sensitive to $h$ and $\alpha$ than ${\bf p}_1$, therefore, ${\bf p}_1$ can be assumed to be unaffected by error in $h$ and $\alpha$. In particular, if we constrain the trained function to always output a set $(t_H, t_L)$ for which point ${\bf p}_2$ is inside a ball of radius $\epsilon$, it must satisfy $|h\Delta \alpha|< \epsilon$ and $|\Delta h|< \epsilon$, assuming $\Delta \alpha$ to be small. An estimate of the relation between $\Delta h$ and $\Delta t_L$ is $\Delta h \sim v_{\omega_L} \Delta t_L$. A linear fitting can also be used to obtain an estimate to $\Delta \alpha \sim c\Delta t_H$. Therefore, the error constraints can be written in the form of $|h c\Delta t_H|< \epsilon$ and $| v_{\omega_L}\Delta t_L|< \epsilon$, implying that $h ct_H$ and $v_{\omega_L}t_L$ must be trained instead of $t_H$ and $t_L$. We have tested this training process and have observed it gives an error similar to that of separate training of $f_L$ and $f_H$, while parameterization based on $t_H$ and $t_L$ is highly biased toward $t_L$ (due to $h c/v_{\omega_L} \gg 1$).

\end{document}